\documentclass[10pt,a4paper,twocolumn]{article}

\usepackage[left=1.5cm,right=1.5cm,top=2.0cm,bottom=2.0cm,columnsep=0.7cm]{geometry}
\usepackage[utf8]{inputenc}
\usepackage[T1]{fontenc}
\usepackage{lmodern}
\usepackage{microtype}
\usepackage{setspace}

\usepackage{amsmath,amssymb,amsthm}
\usepackage{bm}
\usepackage{algorithm}

\usepackage{graphicx}
\usepackage{tikz}
\usetikzlibrary{arrows.meta,positioning,shapes.geometric,fit,backgrounds}
\usepackage{booktabs}
\usepackage{enumitem}
\usepackage{xcolor}
\definecolor{accent}{RGB}{0,80,158}
\definecolor{accentlight}{RGB}{231,241,250}
\definecolor{accentdark}{RGB}{20,40,90}
\definecolor{soft}{RGB}{120,130,145}

\usepackage{titlesec}
\titleformat{\section}{\large\bfseries\color{accentdark}}{\thesection}{0.5em}{}
\titleformat{\subsection}{\normalsize\bfseries\color{accent}}{\thesubsection}{0.45em}{}
\titleformat{\subsubsection}{\small\bfseries}{\thesubsubsection}{0.4em}{}

\titlespacing*{\section}{0pt}{8pt plus 1pt minus 1pt}{3pt plus 0pt minus 0pt}
\titlespacing*{\subsection}{0pt}{6pt plus 1pt minus 1pt}{2pt plus 0pt minus 0pt}
\titlespacing*{\subsubsection}{0pt}{4pt plus 1pt minus 0pt}{1pt plus 0pt minus 0pt}

\newcommand{\mypara}[1]{\vspace{1mm}\noindent\textbf{#1}}

\usepackage[numbers,sort&compress,square,comma]{natbib}


\usepackage{balance}

\usepackage{hyperref}
\hypersetup{
    colorlinks=true,
    linkcolor=accent,
    citecolor=accent,
    urlcolor=accent,
    pdftitle={Cohort-Anchored Foundation Models for EHR},
    pdfauthor={Kaiping Zheng}
}

\newcommand{\name}[1]{\textsc{#1}}
\usepackage{xspace}
\newcommand{\cafm}{\name{cafm}\xspace}

\setlist{topsep=1pt,itemsep=1pt,partopsep=0pt,parsep=0pt}

\begin{document}

\twocolumn[
\begin{@twocolumnfalse}
  \begin{center}
    {\LARGE\bfseries\color{accentdark}
      Cohort-Anchored Foundation Models for Electronic Health Records: From Risk Scores to Auditable Peer Cohorts\par}
    \vspace{14pt}
    {\normalsize Kaiping Zheng\par}
    \vspace{2pt}
    {\small\itshape\color{soft}
      National University of Singapore\par}
    \vspace{2pt}
    {\small\texttt{dcszkai@nus.edu.sg}\par}
    \vspace{14pt}
  \end{center}
  \begin{center}
\begin{minipage}{0.94\textwidth}
\noindent\textbf{Abstract.}\quad
Foundation models are reshaping clinical artificial intelligence, with
billion-parameter language and vision backbones demonstrating impressive zero-
and few-shot abilities on medical question answering, imaging, and structured
electronic health record (EHR) tasks. Yet the gap between benchmark accuracy
and reliable clinical deployment remains stubbornly wide: predictions are
opaque, brittle under distribution shift, and disconnected from the way
clinicians actually reason, through \emph{cohorts} of similar patients with
documented outcomes. The gap is structural: the prevailing recipe optimises
for representation quality and treats the patient comparison as an emergent
side-effect rather than as the primary unit of clinical evidence.
To this end, we propose \cafm, a
\textbf{C}ohort-\textbf{A}nchored \textbf{F}oundation \textbf{M}odel framework
that elevates the cohort to a first-class object in four interlocking stages:
deviation-aware data curation that surfaces low-quality or biased records
before pretraining; cohort-conditioned pretraining that organises the latent
space along clinically meaningful axes via a contrastive objective with
near-miss negatives; multimodal cohort alignment that resists representation
contamination by preserving intra-cohort similarity within each modality; and
clinician-in-the-loop refinement that translates model behaviour into
auditable evidence and a deployment drift signal. The framework is
compositional: existing EHR backbones can be retrofitted with cohort-aware capabilities without rebuilding the encoder. We illustrate the \cafm framework
through four case studies (acute kidney injury prediction from longitudinal
laboratory tests, cardiovascular risk stratification from electrocardiograms,
optic neuropathy triage from orbital imaging, and electroretinogram-grounded
report generation) drawn from operational settings, articulate five empirical
hypotheses that the framework predicts and that could refute it, and conclude
with an agenda of open problems spanning data quality, irregular temporality,
multimodal collapse, distribution shift, and evaluation beyond accuracy.
Explicitly anchoring foundation models to patient cohorts is, in our view,
the most direct route from impressive benchmarks to trustworthy bedside
use.
\end{minipage}
\end{center}

  \vspace{12pt}
\end{@twocolumnfalse}
]

\section{Introduction}
\label{sec:introduction}

\begin{figure*}[t]
\centering
\includegraphics[width=.95\linewidth]{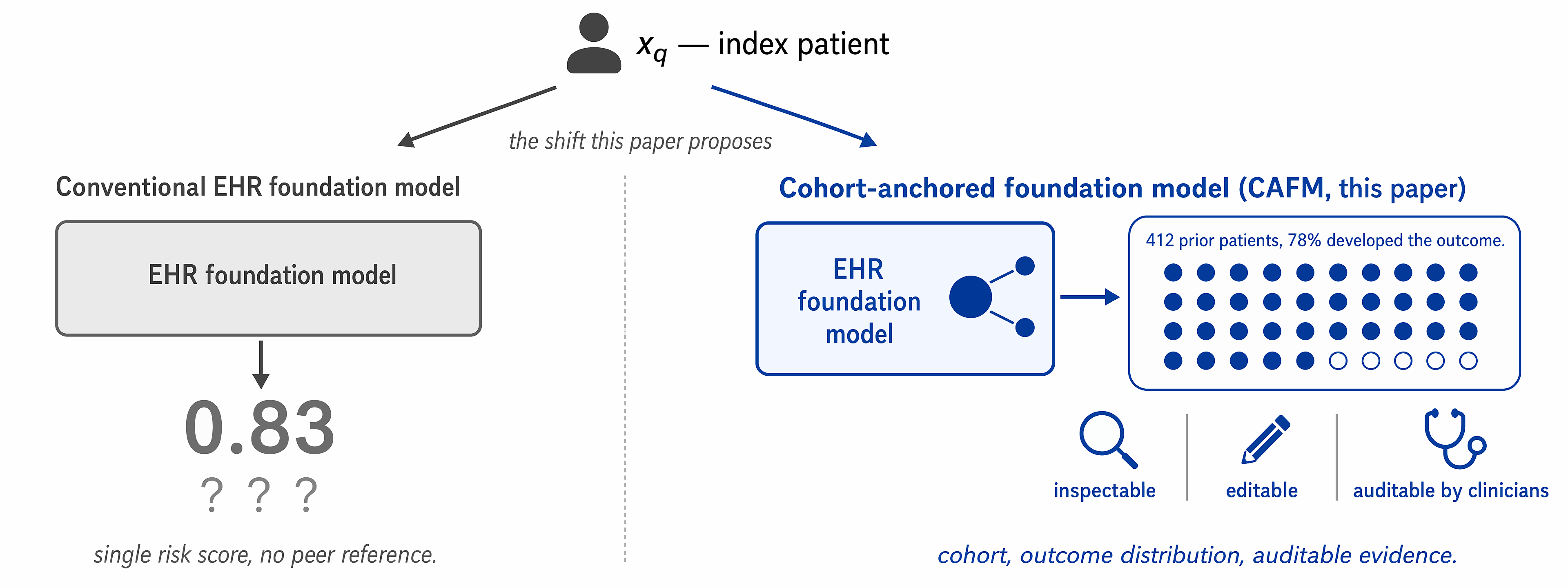}
\vspace{-3mm}
\caption{
The proposed shift: the same backbone, but a different output unit. Instead of producing a scalar risk score, the model returns a peer cohort and its associated outcome distribution, which jointly serve as the explanation, audit unit, and drift signal.}
\label{fig:teaser}
\end{figure*}

The promise of artificial intelligence in medicine is, at heart, the promise of
\emph{learning from many patients to help one}. Clinicians already do this:
rounding through wards, recalling similar cases from training and prior
consultations, benchmarking a new admission against the peer group it most
resembles, and committing
to a plan that is, in the end, defensible because peers like this one have responded
to it before. For decades, machine learning systems trained on electronic health
records (EHRs) have approximated this process indirectly: the patient's history is
mapped through a discriminative classifier to a scalar risk
score~\citep{rajkomar2018scalable, choi2016retain, choi2016doctorai,
shickel2018deepehr}. However, the resemblance to clinical reasoning has always been
superficial. The classifier sees a patient as a point in a latent space and produces
a probability; the clinician sees a patient as a member of a cohort and produces a
plan. Bridging this gap is the central goal of the paper, and
Figure~\ref{fig:teaser} provides an overview of the proposed paradigm shift.

The recent rise of foundation models has changed the surface of the problem without,
we will argue, changing its core. The same self-supervised recipe that delivered
breakthroughs in language~\citep{vaswani2017attention, devlin2019bert,
brown2020gpt3, touvron2023llama, openai2023gpt4} and vision~\citep{radford2021clip,
oquab2024dinov2} has produced clinical and biomedical
backbones~\citep{singhal2023medpalm, singhal2025medpalm2, moor2023foundation,
tu2024towards, yang2022gatortron, lu2024conch, mckeen2024ecgfm, krishnan2022ssl}
that, with minimal task-specific supervision, transfer well across questions ranging
from medical examinations to imaging triage to physiological signal interpretation.
A single backbone now plays the role that a fleet of bespoke classifiers
used to play, and the economics of clinical machine learning have improved
correspondingly. Yet the structural mismatch is unchanged: the model still
produces a representation and a score, while the clinician still needs a
comparison group with a documented trajectory.

The gap between benchmarks and the bedside manifests in three specific
failure modes, each well documented in the
literature and already visible in current systems.
First, foundation models trained on web-scale text excel at examination-style
question answering~\citep{singhal2023medpalm, nori2023capabilities, lievin2024canllm}
but are known to hallucinate plausible-yet-incorrect medical content with
confidence~\citep{ji2023survey, umapathi2023medhalt}. The hallucinations are
particularly damaging in clinical contexts because surface fluency masks the
substantive error: an answer that reads like a cardiology consultation can be incorrect
in precisely the details that a real consultation would identify.
Second, predictive performance degrades sharply under domain shift. Pneumonia
detectors trained on one hospital's chest radiographs fail to generalise to
another~\citep{zech2018variable}; clinical risk models drift across patient
demographics, sites, and protocols~\citep{finlayson2021clinician,
yang2024demographic, futoma2020myth}; and the behaviour of structured EHR encoders
on novel patient subpopulations is poorly characterised. The pattern is consistent
enough to be considered a property of the recipe, not a series of unfortunate
implementation choices.
Third, the explanations these models produce do not survive bedside
scrutiny~\citep{ghassemi2021false, rudin2019stop}. Saliency maps, attention weights,
and feature attributions answer questions clinicians did not ask: ``which input
tokens did the model weigh?'' rather than ``which prior patients support this
decision, and what were their clinical outcomes?'' The structured-EHR side of the field fares no
better in this regard. Encoders such as BEHRT~\citep{li2020behrt},
Med-BERT~\citep{rasmy2021medbert}, CEHR-BERT~\citep{pang2021cehrbert},
and MOTOR~\citep{steinberg2024motor} achieve state-of-the-art predictive
performance, yet their outputs are still risk scores over predefined outcomes, rather than the cohort-grounded justification that a clinician would document in the medical record. A recent review~\citep{wornow2023shaky} catalogues 84 such models
and concludes that the majority are evaluated on tasks that do not provide
meaningful insight into their usefulness to health systems; benchmarks such as
EHRSHOT~\citep{wornow2023ehrshot}, MedHELM~\citep{bedi2025medhelm}, and
MedMCQA~\citep{pal2022medmcqa} sharpen the picture in successive
generations: on tasks where labels are scarce or where patient subgroups behave
heterogeneously, the advantage of pretraining shrinks, and reliability concerns
grow.

Why has the surface progress not closed the gap? Because the prevailing recipe
optimises for \emph{representation quality}, that is, how informative the
latent embedding is when probed by a classifier, and treats patient
comparison as an emergent side-effect of that representation. The clinician, by contrast, does not reason over an
embedding; she reasons over a cohort. A cohort is a structured set of clinically
similar patients, defined explicitly enough that membership can be evaluated for any candidate patient. 
Unlike embeddings, cohorts support clinically meaningful operations such as
counting, subgroup analysis, conditioning on outcomes, removing contaminated members, and
mapping back to established phenotyping ontologies. Cohort-based reasoning is the lingua franca of
evidence-based medicine~\citep{hripcsak2013phenotyping, denny2010phewas,
hripcsak2015observational}: clinical trials are designed around cohorts, guidelines are
written for cohorts, and treatment decisions are routinely framed in terms of ``patients like this
one tend to ...''.
The cohort is the unit of evidence because it is also the unit of clinical accountability.

Recent work has begun to bring cohort-style reasoning into deep learning, but the
threads have not been integrated with the foundation-model paradigm.
\emph{CohortNet}~\citep{cai2024cohortnet} treats cohort discovery as a first-class
learnable component rather than a downstream clustering pass.
\emph{NeuralCohort}~\citep{liu2025neuralcohort} shows that cohort-aware
representations transfer across heterogeneous downstream tasks and task-specific encoders, validating the population-grounded view at scale. Even the
choice of which patients \emph{not} to include shapes what the model
learns: \citet{zheng2024exploiting} demonstrate that carefully curated negative
samples are a catalyst for sharper cohort boundaries, with the gain coming not from
more data but from harder negatives. 
Beyond cohort discovery itself, 
the field has begun to retrieve at inference time (MedRAG~\citep{xiong2024medrag}
and RAM-EHR~\citep{xu2024ramehr}), but the retrieved unit is a document or
knowledge snippet, not a patient cohort. 
Cohort anchoring requires a stronger structural commitment: retrieval over clinically comparable \emph{patients} themselves. It is precisely this patient-centered retrieval substrate that current foundation-model pipelines continue to lack.

This paper presents a perspective backed by a concrete framework. Our
contributions are:
\begin{itemize}[leftmargin=1.5em,topsep=2pt,itemsep=2pt]
\item We argue that \emph{cohort anchoring} is a structural
prerequisite for EHR foundation models, distinct from in-context
learning, retrieval-augmented generation, post-hoc explanation,
prototype networks, and patient-level k-nearest neighbors (kNN) over learned embeddings;
Section~\ref{sec:discussion} develops these distinctions precisely
rather than rhetorically.
\item We propose \cafm, a four-stage framework (deviation-aware cohort
construction, cohort-conditioned pretraining, multimodal cohort
alignment, and clinician-in-the-loop refinement) and connect each stage to the existing
primitives in the literature. The \cafm framework is intended as a
\emph{retrofit} on top of an existing EHR backbone rather than a
replacement; the cost of adoption is engineering effort rather than a
new pretraining run. We accompany the framework with explicit
mathematical formulations of the contrastive objective, an
implementation sketch (Algorithm~\ref{alg:cafm}), and a discussion of several involved issues
that would otherwise compromise the joint training of cohort modules and
encoders.
\item We surface five open challenges (data deviations, irregular
temporality, modality contamination, distribution shift, and
evaluation beyond accuracy) that the framework brings into sharper
focus, and indicate concrete research questions for each
(Section~\ref{sec:challenges}).
\item We illustrate the framework through four case studies (acute
kidney injury (AKI) prediction, cardiovascular risk stratification from
electrocardiograms (ECG), optic neuropathy triage from orbital imaging, and
electroretinogram (ERG)-grounded report generation) drawn from operational
settings, and 
discuss the benefits that cohort anchoring provides in each setting, as well as the associated trade-offs and costs (Section~\ref{sec:case_studies}).
\item We state five \emph{empirical hypotheses}
(Section~\ref{sec:hypotheses}) implied by the framework, each of which could be empirically tested by a single research group within a relatively short time frame. 
We further provide a detailed \emph{scope-and-caveats} discussion (Section~\ref{sec:limitations}) that identifies the boundary conditions and deployment assumptions that practitioners should evaluate before adopting the framework.
\end{itemize}
We frame the work as a \emph{perspective with a concrete framework}, following the
tradition of recent agenda-setting studies in clinical
AI~\citep{moor2023foundation, acosta2022multimodal, rajpurkar2022aiopia,
topol2019highperformance, thirunavukarasu2023llm}, while placing stronger emphasis on methodological specificity and implementation guidance. 
The intended reader is the practitioner initiating a clinical foundation-model project who is willing to invest additional
engineering effort to make the resulting system interpretable to clinicians,
robust under shift, and auditable in real-world deployment. 
Our central argument is that cohort anchoring provides a favourable trade-off between this additional engineering complexity and the resulting gains in these properties.

\section{Background and Motivation}
\label{sec:background}

\subsection{The rise of healthcare foundation models}

The current generation of healthcare foundation models can be broadly grouped into four families, mirroring the historical progression of foundation models in machine learning: from language models to vision models, to multimodal models, and ultimately to physiological-signal models.
Each family has inherited the strengths and the structural limitations of its general-domain ancestor.

\mypara{Clinical and biomedical language models} began as in-domain extensions of
BERT~\citep{devlin2019bert}: Clinical BERT~\citep{alsentzer2019clinicalbert}
fine-tuned on MIMIC discharge notes, BioBERT~\citep{lee2020biobert} on
PubMed abstracts and PubMedBERT~\citep{gu2021pubmedbert} on biomedical
text from scratch. The recipe scaled with the language model
revolution: BioGPT~\citep{luo2022biogpt} brought autoregressive generation
to biomedicine, GatorTron~\citep{yang2022gatortron} pretrained on tens of
billions of clinical-text words at a single health system, and the
Med-PaLM family~\citep{singhal2023medpalm, singhal2025medpalm2} achieved
expert-level performance on USMLE-style questions. Reasoning-tuned and
multimodal-aware variants followed quickly~\citep{nori2023capabilities,
saab2024medgemini, lievin2024canllm}. This progression has been impressive on
benchmark evaluations, yet two empirical findings complicate this
otherwise encouraging trajectory.
\citet{lehman2023clinicalLM} show that for many EHR-specific tasks,
specialised clinical models still outperform much larger general-purpose ones in
the few-shot regime, suggesting that scale does not compensate for distributional
mismatch; and~\citet{thirunavukarasu2023llm} argue that the pace of capability
gains is outstripping the pace of evaluation, leaving deployers with
limited solid ground on which to base their trust in these systems.

\mypara{Structured EHR encoders} take a different shape, more interesting for our
purposes. Rather than tokenising free text, they tokenise the longitudinal
record itself (ICD codes, lab values, medication orders, vitals) into event
sequences and apply masked or autoregressive pretraining over them.
BEHRT~\citep{li2020behrt} pioneered the recipe;
Med-BERT~\citep{rasmy2021medbert} scaled it to 28 million patients;
CEHR-BERT~\citep{pang2021cehrbert} added explicit time encoding;
MOTOR~\citep{steinberg2024motor} reframed the objective as time-to-event
prediction across many outcomes simultaneously; and the most recent
CEHR-GPT~\citep{pang2024cehrgpt} demonstrated that the chronological event
stream can be generatively modelled with sufficient fidelity to support synthetic
EHR generation. These models have absorbed the methodology of language modelling
almost wholesale, and they inherit the same limitation: their output is a
sequence-level probability rather than a grounding in any actual
patient population.

\mypara{Medical vision-language models} extend the recipe to imaging.
CONCH~\citep{lu2024conch} aligns histopathology slides with diagnostic
reports; BiomedCLIP~\citep{zhang2023biomedclip} performs the same alignment
at scale across image-caption pairs from biomedical articles;
PLIP~\citep{huang2023plip} mines pathology Twitter for an unconventional
but informative training corpus; and
Med-Flamingo~\citep{moor2023medflamingo} integrates the two modalities
through few-shot learning. 
The clinical promise is real, but the
contamination problem discussed further in Section~\ref{sec:framework:multimodal} is particularly evident in this setting. 
When a pretrained language model dominates a comparatively shallow image encoder, the resulting multimodal model can behave more like a sophisticated caption generator than a robust diagnostic system, leading to reduced discriminative performance under previously unseen image distributions.

\mypara{Physiological-signal foundation models} are the youngest of the four
families and the most directly relevant to bedside applications.
ECG-FM~\citep{mckeen2024ecgfm} pretrains transformer encoders on millions
of ECG tracings, while ST-MEM~\citep{na2024stmem} introduces spatio-temporal masking
tailored to the 12-lead structure.
PTB-XL-trained
benchmarks~\citep{strodthoff2021ptbxl} 
provide reproducible evaluation protocols for this emerging area
and \citet{narayanswamy2024scaling} further extend this scaling paradigm to wearable sensor data.
The clinical applicability of these models is
supported by deployment-oriented studies demonstrating that AI-enabled ECG systems can detect left-ventricular
dysfunction~\citep{attia2019artificial} and atrial
fibrillation~\citep{attia2019afib}.
Beyond clinical environments, DIYHealth Suite~\citep{liu2026diyhealth} 
highlights the growing importance of physiological-signal-driven AI for home-based health management, supporting real-world applications such as sleep and cardiovascular monitoring.

These four families share a common methodological backbone, namely masked
or autoregressive self-supervised pretraining on large unlabelled corpora,
followed by adaptation to task-specific labels~\citep{krishnan2022ssl}, and an emerging
ecosystem of evaluation suites~\citep{wornow2023ehrshot, bedi2025medhelm,
pal2022medmcqa}. They have collectively shifted the field from \emph{train one
model per task} to \emph{adapt one backbone to many tasks}, mirroring the general
trajectory of foundation models elsewhere~\citep{bommasani2021foundation}. 
They also share a common limitation. Across all four families, inference is performed by representing the patient as a point in a latent space, and evaluation is framed in terms of the performance score on a predefined task. 
Neither corresponds to the level at which clinicians typically reason about individual patients in practice.

\subsection{The cohort paradigm in clinical practice}

To see why this limitation matters, it helps to look at what clinicians actually do
when they reason under uncertainty. The mode is closer to case-based reasoning
than to logistic regression. A nephrologist seeing a hospitalised patient with a
rising creatinine does not ask ``what is the probability of AKI (short for acute kidney injury)?'' as a stand-alone numeric question; she asks ``among my prior patients
with similar comorbidities, baseline kidney function, and current trajectory, who
developed AKI, and why?'' The answer is therefore structurally a cohort-with-outcome, from which subsequent clinical actions (e.g., fluid challenge, medication adjustment, or nephrology
referral) can be derived through direct inspection. 
This cohort-centered representation makes the decision process auditable both to the treating clinician and to downstream reviewers of the case, because it exposes not merely the model's prediction, but also the evidentiary basis underlying that prediction.

This pattern is so deeply embedded in evidence-based medicine that the field's
infrastructure has been built to support it. Clinical trials are constructed
around cohorts with explicit inclusion and exclusion
criteria~\citep{collins2015tripod, collins2024tripod}; observational studies use
the cohort as the natural unit of confounding adjustment; phenotyping pipelines
such as PheWAS~\citep{denny2010phewas} translate ICD codes 
into reusable cohort definitions; and the OHDSI common data
model~\citep{hripcsak2015observational} exists in part to make those definitions
portable across institutions. Reporting standards
such as TRIPOD~\citep{collins2015tripod}, TRIPOD+AI~\citep{collins2024tripod},
and the recent LLM-oriented extension TRIPOD-LLM~\citep{gallifant2025tripodllm} require
that a clinical prediction model explicitly specify the population on which it was developed; this population corresponds to the cohort.
In short, when medicine wants to reason responsibly
about an individual, it does so by attaching the individual to a cohort.
Some EHR foundation models partially move in this direction.
MOTOR's
time-to-event prediction~\citep{steinberg2024motor} is implicitly population-aware,
and ontology-aware encoders such as GRAM~\citep{choi2017gram} pull
similar phenotypes together. However, none of them returns a cohort-with-outcome as
the inference output, which is the property that makes a model auditable in
the way that established clinical infrastructure has come to expect.

In machine learning, cohort-style reasoning has been approached, but
never elevated to a first-class output. \citet{baytas2017tlstm} subtype
patients with time-aware LSTMs; \citet{suresh2018learning} treat
heterogeneous ICU populations as separate multi-task heads to surface
population heterogeneity; \citet{ming2020protosteer} learn a small set of exemplar cases (i.e., prototypes) that explain a sequence model's predictions to users. \citet{choi2017gram} embed concepts into a medical
ontology so
that rare phenotypes inherit information from their
ancestors. \citet{choi2018mime} go further, modelling the multilevel
structure of EHR visits with auxiliary self-supervised objectives that
respect the visit--code hierarchy.
\citet{halpern2016electronic} leverage a small set of clinician-specified anchors as initialization signals for semi-supervised phenotype learning.
Most recently, end-to-end neural networks for cohort modeling~\citep{cai2024cohortnet, liu2025neuralcohort, zheng2024exploiting}
have shown that the cohort module can be trained jointly with the
predictive model, yielding interpretable groupings without sacrificing
accuracy. 
What remains missing is a principled integration with the foundation-model paradigm that now dominates clinical AI. None of these approaches has been incorporated into the self-supervised pretraining recipe at the corpus scale where such inductive structure becomes most consequential.

\subsection{Why foundation models miss the cohort}

A pretrained EHR encoder produces a sequence of token embeddings; downstream tasks
read a probability off a classification or generative head. Even when the encoder
implicitly groups similar patients in the latent space (and it does, as a
side-effect of the contrastive or masked objective~\citep{chen2020simclr,
he2020moco, caron2021dino}), the grouping is not exposed, not aligned to
clinically meaningful axes, and not compatible with the data-quality discipline
that medicine requires. We unpack three specific obstacles in the
remainder of this section.

First, \emph{cohorts are not exposed}. The model's nearest neighbours in latent
space are accessible only to those with direct access to the retrieval index. As a result, they are unnamed (lacking clinically interpretable labels), unsized (without local density of the
neighbourhood), and unaudited (without mechanisms that allow clinicians to include or exclude members).
Recent work on retrieval-augmented prediction in
medicine~\citep{xiong2024medrag, xu2024ramehr}
mitigates this for text-centric outputs by retrieving evidence snippets at query
time, but no analogous mechanism is standard for structured EHRs, and none
retrieves \emph{patients} as the unit of evidence. The distinction is not
cosmetic: patients are structured objects that can be stratified by outcome,
demographics, temporal window, treatments, and other clinically meaningful attributes, whereas documents can be retrieved and ranked
but not partitioned along them.

Second, \emph{similarity is monolithic}. A single embedding conflates demographics,
laboratory measurements, medications, vital signs, and free-text clinical notes. In practice, however, the cohort most informative for predicting AKI in a 75-year-old patient receiving diuretics is not necessarily the cohort most informative for predicting heart failure in that same patient, despite the underlying individual being identical. 
Prior work on adaptive interaction modelling~\citep{cai2022elda, cai2021armnet} and dynamic networks~\citep{zheng2022dyhealth} partially addresses this issue at the predictor level by routing patients through task-specific substructures, while progression-aware and temporally adaptive encoders~\citep{bai2018interpretable, ma2017dipole, zheng2017capturing} tailor similarity to disease dynamics over time.
Yet these mechanisms remain largely absent from the pretraining objective itself.
Consequently, the resulting backbone inherits a fixed notion of patient similarity determined during pretraining, even though that notion may be poorly aligned with many downstream clinical tasks encountered during real-world deployment.

Third, \emph{the data are not clean}. EHRs are notorious for non-random
missingness, documentation drift, and selection
bias~\citep{weiskopf2013methods, gianfrancesco2018potential, obermeyer2019dissecting}.
The most consequential issues are not noise but structure: a laboratory test is absent because the clinician considered it unnecessary, which itself encodes an implicit clinical
judgment; a diagnosis is recorded only when required for billing, which biases
prevalence by payer systems; a trial-derived risk model is silently more accurate for the
demographic groups most represented among trial participants. 
Pretraining directly on raw EHR streams without explicitly modelling or exposing these biases
risks burning them into the foundation model itself, 
where they are subsequently inherited by downstream cohort retrieval mechanisms~\citep{zheng2025detecting, zheng2017resolving, wornow2023shaky}.
A model that retrieves cohorts without curating them at the source faithfully reproduces historical documentation practices as if they were causal clinical truth.

The \cafm framework proposed in this paper addresses these three challenges directly and in a deliberate sequence: curation first (Section~\ref{sec:framework}, Stage~1), followed by pretraining that incorporates cohort structure (Stage~2), alignment that preserves
modality structure (Stage~3), and finally refinement that reflects clinical workflow (Stage~4).
The ordering is intentional rather than arbitrary: failure modes introduced at one stage propagate into the next, making attempts to address downstream issues without resolving upstream problems largely ineffective.

\section{The CAFM Framework}
\label{sec:framework}

\begin{figure*}[t]
\centering
\includegraphics[width=\linewidth]{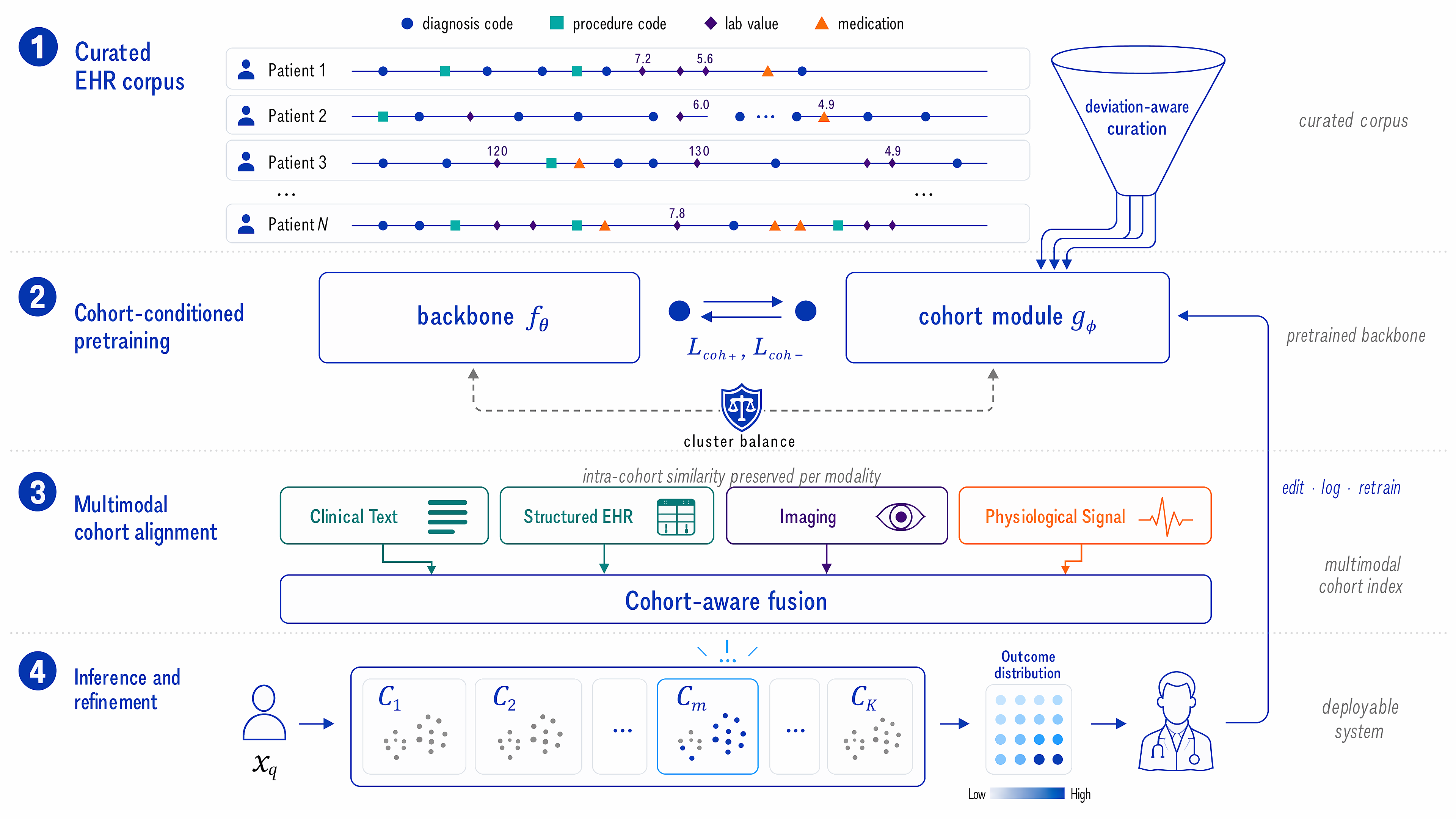}
\caption{The \cafm framework. Four stages
((1) deviation-aware cohort construction, (2) cohort-conditioned
pretraining, (3) multimodal cohort alignment,
(4) clinician-in-the-loop refinement) expose cohorts as
first-class objects.}
\label{fig:framework}
\end{figure*}

\cafm consists of four interlocking stages, summarised in Figure~\ref{fig:framework}.
We describe each in turn, identify the methodological ingredients available today,
and flag where new research is most needed. The stages are designed to be
\emph{compositional}: an existing EHR foundation model can be retrofitted with
cohort anchoring at one or more stages rather than rebuilt from scratch. The cost of
asking a research community to abandon its pretrained backbones is high; the cost of
asking it to add a cohort module and a curated retrieval index is much lower, and we
argue throughout that the latter recovers most of the benefit.

\subsection{Definitions}
\label{sec:framework:definitions}

Before describing the stages, we fix three terms that the rest of the paper uses
repeatedly. Each is given a single, explicit definition to avoid the slippage
between ``cohort as cluster'', ``cohort as soft assignment'', and ``cohort as
retrieved set'' that informal writing about patient similarity tends to suffer
from.

\mypara{Cohort.} A cohort is a (possibly soft) subset of the curated training
corpus, identified by a clinically interpretable handle: a phenotype label, a
prototype index, or a learned categorical variable. Concretely, we represent a
cohort as a discrete index $c \in \mathcal{C} = \{1, \dots, K\}$ together with
the set of training patients whose cohort assignment places them in $c$. A
patient's relationship to a cohort is described by a probability rather than a
hard membership, but the cohort itself is discrete and addressable; this is the
property that distinguishes a cohort from a latent neighbourhood.

\mypara{Cohort module.} The cohort module $g_\phi : \mathcal{X} \to
\Delta^{K-1}$ maps a patient $x$ to a categorical distribution $\pi(c \mid x)
= g_\phi(x)$ over cohorts. The module is trained jointly with the backbone
encoder $f_\theta$; its parameters $\phi$ are updated by gradients flowing
from both the contrastive objective (Section~\ref{sec:framework:pretraining}) and
the downstream task heads. We treat $g_\phi$ as soft throughout, both because
clinical boundaries are fuzzy and because hard assignments preclude the
gradient signal that joint training requires.

\mypara{Cohort anchoring.} A foundation-model pipeline is cohort-anchored
if (i) it includes an explicit cohort module $g_\phi$; (ii) at inference time, it returns a retrieved cohort and a within-cohort outcome distribution
alongside (or in place of) a scalar score; and (iii) the cohort assignments
are exposed for audit, editing, and drift monitoring. The third property
distinguishes cohort anchoring from any pipeline that happens to compute a
nearest-neighbour set internally without surfacing it.

\subsection{Stage 1: Deviation-aware cohort construction}
\label{sec:framework:stage1}

Before any pretraining, the corpus must be vetted. The two failure modes we
have in mind here (\emph{deviation} and \emph{bias}) are easily confused but
mechanistically distinct, and a curation pipeline must address each on its own
terms.

\mypara{Deviation detection.}
Deviation is what one notices when a record's distribution departs from the
population in ways that are not clinically explained: an outlier lab value
that is the legacy of a unit-conversion bug; a vital-sign sequence that is
identical across thousands of patients because the bedside monitor failed
silently; a coded diagnosis that appears only on the day of admission because
of a billing-driven batch upload. Each of these patterns is statistically
detectable and clinically spurious; if left unaddressed in the training corpus, they harden into the
model's prior. We have recently addressed this problem by formalising
deviation as a learnable signal that can be evaluated against held-out
reference cohorts~\citep{zheng2025detecting}, turning curation from a
hand-coded preprocessing pass into a trainable component that can be re-run
as the corpus evolves. The advantage of the trainable view is that it scales:
rather than maintaining a finite set of validity rules per site, the
deviation detector learns the joint distribution of clinically valid records
and reports anything outside the support, allowing curation to keep pace with
documentation practice.

\mypara{Bias resolution.}
Bias is what one notices when a record's distribution is internally
consistent, but the consistency itself is the artefact of a clinical workflow
rather than biology. The classic example is informative missingness: a lab
is missing not at random but because the clinician judged it unnecessary,
which is itself a clinical opinion that the model should not silently absorb
as ground truth~\citep{zheng2017resolving}. The broader literature on EHR
bias documents the consequences of ignoring such structure. Race, sex, and
socioeconomic factors shape what is recorded, when, and how
accurately~\citep{obermeyer2019dissecting, gianfrancesco2018potential,
chen2021ethical, weiskopf2013methods, yang2024demographic}; algorithms
trained on the resulting record reproduce the disparities as predictions,
sometimes
amplified~\citep{obermeyer2019dissecting}. Resolution is rarely reducible to simple imputation. 
A useful rule of thumb is that whenever a missingness mask is
predictive of the outcome, the model should consume the mask explicitly
rather than be asked to ignore it. Counterfactual decision-support
frameworks~\citep{schulam2017reliable} that distinguish ``what was observed''
from ``what would have happened under a different policy'' provide one
principled lens, though their cost (in data assumptions and analytical complexity) is non-trivial.

\mypara{Cohort assembly.}
Once a corpus has been curated for deviation and bias, cohorts can be
assembled along multiple complementary axes. Rule-based phenotypes encode
clinical knowledge directly~\citep{denny2010phewas, hripcsak2015observational}.
Anchor-and-learn phenotyping seeds semi-supervised classifiers from a
small number of clinician-specified anchors~\citep{halpern2016electronic},
providing a human-in-the-loop bridge. Learned
prototypes~\citep{ming2020protosteer} discover groupings without manual
specification. Ontology-guided concept hierarchies~\citep{choi2017gram} embed
cohort structure into the representation space itself. End-to-end neural
cohort networks~\citep{cai2024cohortnet} train the cohort module jointly
with the downstream predictor and produce groupings that are both predictive
and inspectable.
\cafm's commitment is methodological rather than architectural: the cohort
module $g_\phi$ defined in Section~\ref{sec:framework:definitions} is a first-class
component of the model, optimised jointly with the backbone $f_\theta$ and
updated as the corpus evolves. 
This design choice introduces a clear downstream cost: cohort assignments must be computed and stored for every patient during both training and inference. In return, it enables a sequence of benefits that propagate throughout the remaining stages of the framework, including contrastive supervision in Stage~2, modality balancing in Stage~3, and clinician-facing auditability in Stage~4.

\subsection{Stage 2: Cohort-conditioned pretraining}
\phantomsection
\label{sec:framework:pretraining}

\begin{figure*}[t]
\centering
\includegraphics[width=0.74\textwidth]{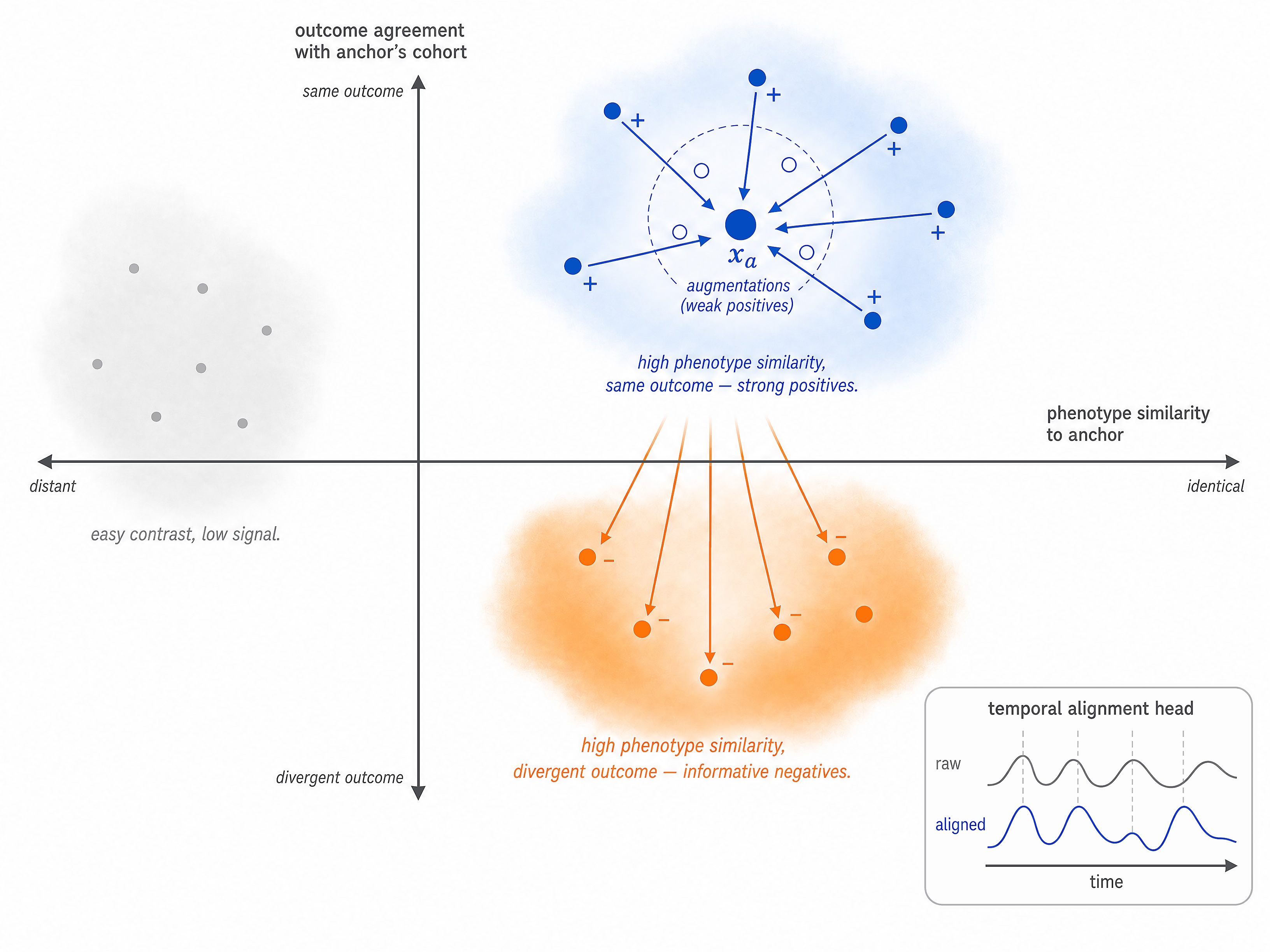}
\caption{Cohort-conditioned pretraining. Anchor $x_a$ pulls toward
augmentation positives and cohort co-members, while pushing away
\emph{near-miss} negatives that share similar phenotypes but differ in outcomes.}
\label{fig:contrastive}
\end{figure*}

Standard EHR pretraining maximises the likelihood of next clinical events
given past ones~\citep{li2020behrt, rasmy2021medbert}.
The objective is sound for next-token modelling, but agnostic to the cohort
structure that downstream tasks ultimately query. \cafm adds a contrastive
objective that pulls representations of patients with similar cohort
distributions together and pushes patients with contrasting cohort
distributions apart, with the goal of organising the latent space along
clinically meaningful axes from the first epoch rather than leaving such
organisation to be discovered post-hoc. The framework borrows from contrastive self-supervised
learning~\citep{chen2020simclr, he2020moco} but adapts
it, as Figure~\ref{fig:contrastive} illustrates, in two specific ways
that match the structure of EHR data.

\mypara{A composite objective.}
Let $f_\theta$ be the backbone encoder, $g_\phi$ the cohort module from
Section~\ref{sec:framework:definitions}, and $z_i = h_\psi(f_\theta(x_i))$ the
projection-head embedding used for contrastive learning. We denote by
$\mathcal{L}_{\text{seq}}$ the standard masked or autoregressive
pretraining objective of an EHR encoder~\citep{li2020behrt, rasmy2021medbert}. 
We define raw cohort similarity between two patients as the inner product of
their soft cohort distributions,
\begin{equation}
s_\pi(x_i, x_j) \;=\; \sum_{c \in \mathcal{C}} \pi(c \mid x_i)\,\pi(c \mid x_j) \in [0,1],
\end{equation}
and use it instead of hard cluster assignments because clinical
boundaries are inherently fuzzy, and because soft-positive relationships permit gradients to propagate through $g_\phi$. To employ $s_\pi$ as the target for a soft-label InfoNCE, we
normalise it over the in-batch comparison set:
\begin{equation}
\tilde{s}_\pi(x_i, x_j) \;=\;
   \frac{s_\pi(x_i, x_j)}{\sum_{x_k \in \mathcal{B}\setminus\{x_i\}} s_\pi(x_i, x_k) + \varepsilon},
\end{equation}
such that $\sum_{x_j \in \mathcal{B}\setminus\{x_i\}} \tilde{s}_\pi(x_i, x_j) = 1$
, thereby yielding a valid cross-entropy interpretation for the objective defined below. 
$\varepsilon > 0$ guards against degenerate batches in which no in-batch peer overlaps with $x_i$. 
We further define the in-batch softmax score
\begin{equation}
p_+(x_i, x_j) \;=\; \frac{\exp(z_i^\top z_j / \tau_+)}
{\sum_{x_k \in \mathcal{B}\setminus\{x_i\}} \exp(z_i^\top z_k / \tau_+)}.
\end{equation}
The intra-cohort attraction term is then formulated as a soft-label InfoNCE, or equivalently,
as a cross-entropy between the normalised cohort target $\tilde{s}_\pi$ and the
softmax distribution $p_+$ computed over the batch:
\begin{equation}
\mathcal{L}_{\text{coh+}}(x_i)
\;=\; -\!\!\!\sum_{x_j \in \mathcal{B}\setminus\{x_i\}}\!\!\!
   \tilde{s}_\pi(x_i, x_j)\,\log p_+(x_i, x_j).
\end{equation}
The near-miss repulsion term penalises representations of patients belonging to cohorts that remain phenotypically similar to the cohort of $x_i$ but differ in outcome. Let $N(x_i) :=
\textsc{NearMissNeg}(x_i)$ denote the corresponding near-miss negative set, defined formally below. The resulting objective is given by
\begin{equation}
\mathcal{L}_{\text{coh-}}(x_i)
\;=\; \frac{1}{|N(x_i)|}\sum_{x_j \in N(x_i)}\!
   \log\!\big(1 + e^{\,z_i^\top z_j / \tau_-}\big).
\end{equation}
To avoid the trivial solution in which $g_\phi$ collapses all patients
into a single cohort assignment (a known failure mode in joint
clustering-and-representation
training~\citep{caron2020swav}), we introduce an entropy-regularisation
or cluster-balance term, $\mathcal{L}_{\text{bal}}$, that encourages the marginal
cohort distribution $\bar{\pi}(c) := \mathbb{E}_x[\pi(c \mid x)]$ to remain close to a
predefined prior over $\mathcal{C}$ (uniform by default).
One simple instantiation is the negative entropy $\mathcal{L}_{\text{bal}} = -H(\bar{\pi})$,
estimated at the batch level.
Combining these terms yields the complete objective used in Stage~2 pretraining:
\begin{equation}
\label{eq:cafm-stage2}
\mathcal{L}_{\textsc{cafm}}
\;=\; \mathcal{L}_{\text{seq}}
\;+\; \lambda_p\,\bar{\mathcal{L}}_{\text{coh+}}
\;+\; \lambda_n\,\bar{\mathcal{L}}_{\text{coh-}}
\;+\; \lambda_b\,\mathcal{L}_{\text{bal}},
\end{equation}
where $\bar{\mathcal{L}}_{\text{coh}\pm}$ denotes the corpus-level expectation of per-anchor losses. Eq.~\ref{eq:cafm-stage2} is further extended in
Section~\ref{sec:framework:multimodal} (Stage~3) with per-modality
intra-cohort terms; the complete pretraining objective is presented in
Eq.~\ref{eq:cafm-full}.
We intentionally separate the attraction and repulsion components rather than coupling them within a single InfoNCE softmax objective for two reasons.
First, this formulation permits the use of distinct temperature parameters $\tau_+$ and $\tau_-$ for the two mechanisms: the attraction term benefits from a sharper softmax to preserve informative gradients within the in-batch comparison pool, whereas the repulsion term benefits from a smoother one to avoid unstable gradients arising from borderline near-misses. 
Second, the decoupled formulation allows the weighting coefficients $\lambda_p$ and $\lambda_n$ to be scheduled independently, both with respect to each other and relative to the balancing coefficient $\lambda_b$.

\mypara{Cold start and curriculum.}
A na\"{\i}ve joint-training strategy fails because, at the beginning of training, the cohort module
$g_\phi$ produces an essentially uniform distribution over $\mathcal{C}$, so
$s_\pi(x_i, x_j) \approx 1/K$ for every pair and the contrastive head
receives no informative gradients through $\pi$. We mitigate this issue through three complementary mechanisms.
First, we \emph{warm-start} $g_\phi$ using a deviation-aware
phenotyping pipeline (Section~\ref{sec:framework:stage1}): a small number of
rule-based or anchor-and-learn cohort
labels~\citep{halpern2016electronic, denny2010phewas, hripcsak2015observational}
provide a non-trivial initial $\pi(c \mid x)$ for a subset of the corpus,
and $g_\phi$ is further pretrained to approximate these labels before contrastive training begins. 
Second, we \emph{ramp} the contrastive coefficients: $\lambda_p,
\lambda_n$ are initially set to small values and increased progressively once $g_\phi$ stabilises, a standard remedy for the collapse issue that sets in when models are forced to commit to noisy positives prematurely~\citep{caron2020swav, caron2021dino}. 
Third, we
\emph{alternate} updates between $g_\phi$ and the contrastive head with a
slow exponential moving average of the cohort module's targets, in the
manner of momentum encoders~\citep{he2020moco, caron2021dino}; this
stabilises the moving-target dynamic that joint training otherwise creates.
Collectively, these mechanisms address the fundamental chicken-and-egg dependency between representation learning and cohort assignment that generic joint training often overlooks. 
While none of the individual components is brand new in isolation, their coordinated integration is essential for stable and effective cohort-anchored pretraining.

\mypara{Defining \textsc{NearMissNeg}.}
The sampler operates over cohort-level statistics that we maintain as moving
quantities throughout pretraining. Let $Y(c)$ denote a designated
\emph{outcome-proxy} distribution within cohort $c$. Two designs are
admissible. In a fully self-supervised pretrain, $Y(c)$ can be the
distribution of \emph{future events}, such as the next $H$ event types in a
patient trajectory, marginalised across the cohort and therefore observable at
training time without external supervision. In a partially supervised setting
with access to inexpensive proxy outcomes (e.g., 30-day readmission derived directly from admission records), $Y(c)$ may instead correspond to the distribution of those proxy labels; in this regime, the line between ``pretraining'' and ``multi-task supervised pretraining'' is blurred.
In either regime, $Y(c)$ is
maintained as an exponential moving average over batches in which cohort
$c$ appears. The phenotype overlap $\rho(c, c')$ is computed offline from
patient-level codes and refreshed periodically. For an anchor patient $x_i$ assigned to cohort $c_i$, the sampler assigns each candidate cohort $c'$ the score
\begin{equation}
\rho(c_i, c') - \beta\,D_{\mathrm{KL}}(Y(c_i) \,\|\, Y(c')), 
\end{equation}
after which cohorts exceeding a predefined quantile threshold are selected, and patients are sampled uniformly from within those selected cohorts.

The sign convention is deliberate: cohorts exhibiting high phenotypic similarity but divergent outcome distributions provide precisely the type of informative contrast that is most valuable for cohort discovery~\citep{zheng2024exploiting}, and the
broader hard-negative-mining literature converges on the same
shape~\citep{robinson2021contrastive, kalantidis2020hard}. The
hyperparameter $\beta$ controls the trade-off between phenotypic similarity and outcome divergence within the near-miss sampling pool. 
Its optimal setting is task- and dataset-dependent, and we therefore treat it as one of the core hyperparameters that any implementation must tune, alongside the contrastive temperatures $\tau_\pm$ and the cohort cardinality $K$.

\mypara{Cohort-aware sampling.}
Augmentations derived from the same patient (e.g., token masking, temporal jittering, or modality
dropout) supply \emph{weak} positives via $s_\pi(x_i, \tilde{x}_i) \approx
1$. Cohort co-membership, in contrast, provides \emph{strong} positives at scale $|\{x_j
: s_\pi(x_i, x_j)>\epsilon\}|$, while the near-miss sampler supplies clinically
informative negative examples. By comparison, 
uniformly sampled random negatives constitute an overly simplistic contrastive signal that the model can separate without learning clinically meaningful cohort structure.
The underlying clinical intuition is concrete: two
patients with highly similar comorbidity profiles and admission characteristics, where
one subsequently develops AKI and the other does not, are precisely the type of informative contrast a clinician would want the model to internalise during pretraining.

\mypara{Time-aware similarity.}
Clinical trajectories are inherently irregular: vital signs are densely sampled, laboratory measurements are sparse, and the timing of events itself often carries substantial clinical information~\citep{shukla2021multitime, tipirneni2022strats, zhang2023warpformer}.
For example, two patients may exhibit nearly identical creatinine trajectories after alignment onto a shared timeline, yet correspond to markedly different clinical conditions depending on whether the increase occurred over several hours or over several weeks.
Prior work on feature-level temporal irregularity~\citep{zheng2017capturing}, dynamic network
adaptation~\citep{zheng2022dyhealth}, reverse-time
attention mechanisms~\citep{choi2016retain}, and continuous-time
architectures~\citep{rubanova2019latent, kidger2020neural} can
be embedded as inductive biases into the contrastive head, so that
cohort similarity is defined not with respect to raw timestamps alone, but relative to clinically meaningful temporal alignment.
A straightforward implementation is a
learned warping head that re-anchors patient timelines onto disease-specific
event clocks (e.g., time since first elevated creatinine) prior to contrastive comparison; more advanced variants further make the warping itself
a function of the cohort assignment. We treat the choice among these
alternatives as an empirical design question.

\subsection{Stage 3: Multimodal cohort alignment}
\label{sec:framework:multimodal}

\begin{figure*}[t]
\centering
\includegraphics[width=0.80\textwidth]{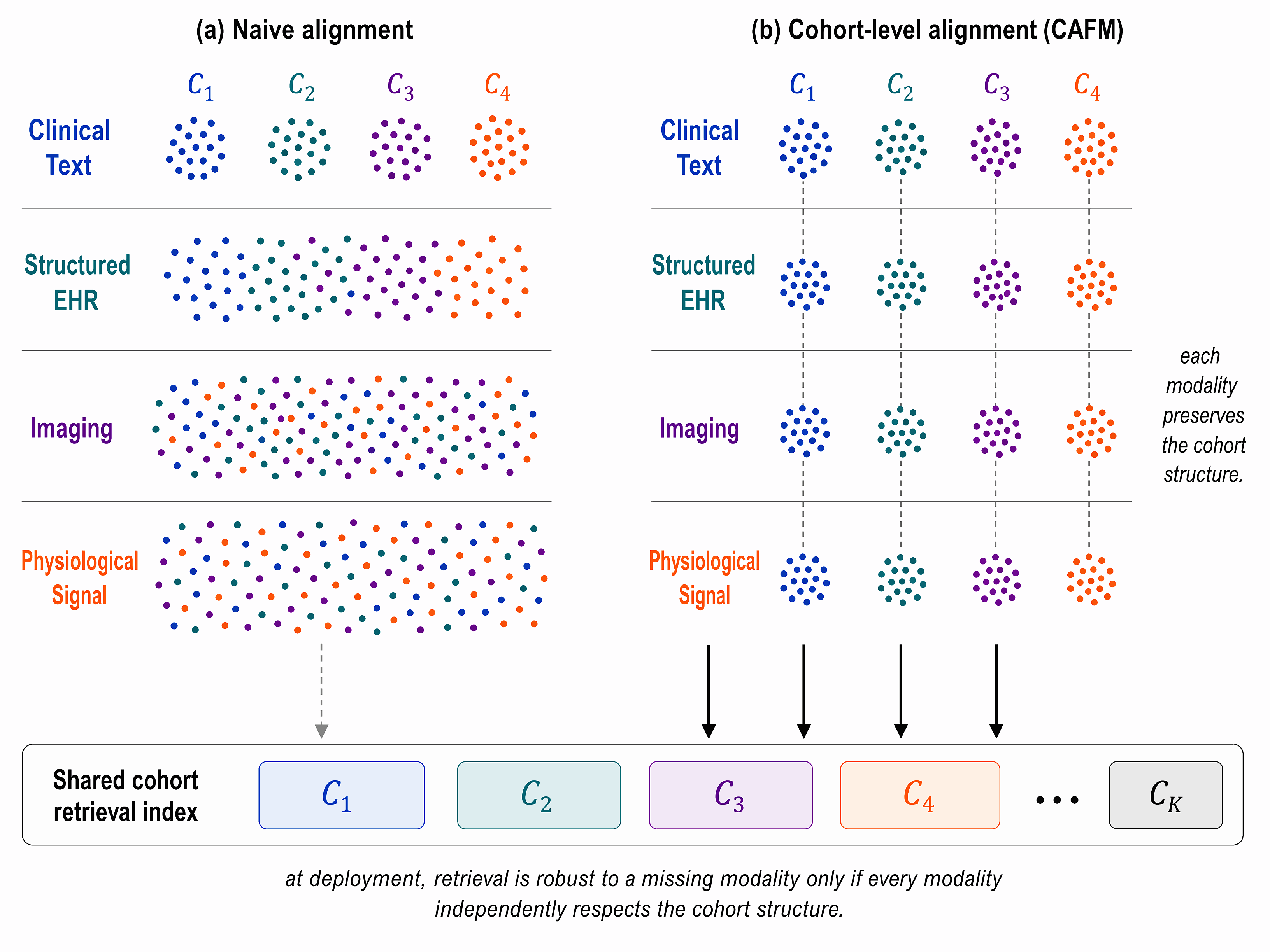}
\vspace{-4mm}
\caption{Multimodal cohort alignment. Per-modality encoders for text,
structured records, imaging, and physiological signals feed into a fusion module that
preserves cohort boundaries. Cohort-level intra-modal preservation mitigates modality dominance, while the cohort retrieval index
supports the clinician-facing audit and retrieval capabilities introduced in Stage~4.}
\label{fig:alignment}
\end{figure*}

EHRs are intrinsically multimodal: structured codes, free-text notes,
imaging, and physiological signals coexist for the same patient and, in the
more ambitious clinical use cases, must inform the same prediction.
Multimodal pretraining in medicine has progressed
rapidly~\citep{acosta2022multimodal, hager2023multimodal, zhang2023biomedclip,
lu2024conch, moor2023medflamingo, huang2023plip}, but these models typically
align modalities \emph{patient by patient}, leaving cohort structure
implicit. The cost of leaving it implicit is two specific failure modes that
\cafm directly addresses as shown in Figure~\ref{fig:alignment}.

\mypara{Representation contamination.}
The first failure mode is contamination. When modalities differ substantially in
representational density (e.g., richly expressive clinical text, intermediate structured-code representations, sparse physiological signals, or highly variable imaging data), na\"{\i}ve multimodal pretraining drives the
joint representation toward whichever modality dominates the optimisation objective. As a consequence, weaker
modalities may collapse onto the dominant one and, in the worst case, become
functionally decorative: gradients continue to propagate through them, but little modality-specific information is retained~\citep{su2025clear, hessel2020emap}. Cohort-level alignment provides a natural mechanism for mitigating this problem. We instantiate it through a per-modality variant of the attraction loss introduced in Section~\ref{sec:framework:pretraining}. For each modality
$m$, let $z_i^{(m)}$ denote the modality-specific projection and $p_+^{(m)}$ denote the corresponding in-batch softmax computed over $\{z_i^{(m)}\}$. For each modality, we then define the corresponding within-modality loss term as
\begin{equation}
\mathcal{L}_{\text{intra}}^{(m)}(x_i)
\;=\; -\!\!\!\sum_{x_j \in \mathcal{B}\setminus\{x_i\}}\!\!\!
   \tilde{s}_\pi(x_i, x_j)\,\log p_+^{(m)}(x_i, x_j),
\end{equation}
which uses the same normalised cohort target $\tilde{s}_\pi$ as the joint
loss but enforces intra-cohort similarity inside each individual modality. Incorporating
these per-modality terms into the Stage~2 objective in
Eq.~\ref{eq:cafm-stage2} yields the complete multimodal pretraining objective:
\begin{equation}
\label{eq:cafm-full}
\begin{aligned}
\mathcal{L}_{\textsc{cafm}}
\;=\;& \mathcal{L}_{\text{seq}}
\;+\; \lambda_p\,\bar{\mathcal{L}}_{\text{coh+}}
\;+\; \lambda_n\,\bar{\mathcal{L}}_{\text{coh-}} \\
&\;+\; \lambda_b\,\mathcal{L}_{\text{bal}}
\;+\; \sum_{m} \lambda_m\,\bar{\mathcal{L}}_{\text{intra}}^{(m)}.
\end{aligned}
\end{equation}
This formulation prevents any single modality from absorbing the others, because
doing so would require that modality's intra-cohort structure to approximate the full joint cohort structure. In practice, this is only achievable when all modalities carry comparable
cohort information, which corresponds to the desired equilibrium. 
The cost is the introduction of one
additional contrastive head per modality.
In return, each modality preserves its own discriminative capacity and can subsequently be evaluated, debugged, and ablated independently. 
Furthermore, gradient-balancing
strategies~\citep{wang2020what, peng2022balanced} and gradient-conflict
projection methods~\citep{yu2020pcgrad} can be integrated with the cohort-level
penalty to maintain calibrated modality contributions during optimisation, 
complementing the architectural safeguards described above with optimisation-level calibration.

\mypara{Cross-domain noise.}
The second failure mode is cross-domain noise. Adapting vision backbones pretrained on natural-image corpora to medical applications (a common practice in
retinal imaging, orbital imaging, and dermatology pipelines) transfers not only useful inductive biases, but also undesirable source-domain noises~\citep{zhu2021robustcrossdomain, ganin2016domain}. 
In the absence of an explicit cohort prior, the model lacks a principled mechanism for down-weighting examples that appear similar in the natural-image sense but are clinically mismatched.
Introducing a cohort prior filters cross-domain adaptation through a clinically grounded similarity structure: adaptation is encouraged when source and target cohorts exhibit meaningful overlap, and constrained when they do not. 
Existing work on cross-domain adaptation~\citep{zhu2021robustcrossdomain} provides suggestive but incomplete support, and controlled cohort-prior ablation studies have not
yet been conducted.

At this stage, the framework produces \emph{multimodal cohort prototypes}:
per-modality encoders, a fusion module that respects cohort
boundaries, and a retrieval index keyed by cohort identity rather than by
raw embedding similarity. The retrieval index supports the clinician-facing
functionalities introduced in Stage~4 and also provides a natural integration point for text-grounded retrieval-augmented generation~\citep{xiong2024medrag, xu2024ramehr} when
free-text clinical justification is required as part of the workflow.

\subsection{Stage 4: Clinician-in-the-loop refinement}
\label{sec:framework:in-the-loop}

A cohort-anchored model exposes capabilities that are unavailable in conventional black-box predictors. Three are particularly important, and each requires substantially more deliberate design than is typically assumed.

\mypara{Auditable evidence.}
Each prediction is accompanied by a retrieved peer cohort together with its associated within-cohort outcome distribution. This transforms a bare prediction such as ``the model outputs 0.83'' into a substantially richer clinical object: ``among 412 prior patients assigned to this cohort, 78\% developed the outcome within 48 hours; the cohort matches the index patient with respect to age, comorbidity profile, and admission trajectory; and a cohort-level summary of prior clinical trajectories is available for clinician inspection at runtime.''
Importantly, the clinician interacts with a structured cohort summary rather than raw patient records. This summary may include cohort-level statistics, the most informative anchor variables, and (subject to institutional policy) a de-identified case vignette synthesised from the cohort. Direct exposure of identifiable peer records would introduce privacy concerns, which we discuss further in Section~\ref{sec:discussion}. Nevertheless, clinicians can still interrogate the cohort in clinically meaningful ways, for example, ``were cohort members exposed to nephrotoxic medications at prediction time?'' or ``what proportion received contrast that day?'' These queries operate within-cohort distributions rather than individual patient charts.
This form of explanation differs fundamentally from feature-attribution methods such as SHAP or LIME~\citep{lundberg2017shap, ribeiro2016should}. Feature attribution identifies which input variables most influenced the prediction, whereas cohort-level provenance identifies the prior patient population that supports the prediction itself. It also generalises beyond what fixed-task interpretable analytics provide~\citep{zheng2020tracer}: cohort-level provenance is meaningful for any new task, because the cohort is a population rather than a fixed task-specific set of explanatory features.

\mypara{Task decomposition with humans in the loop.}
Cohort-anchored systems naturally decompose clinical decision-making into a sequence of sub-decisions, each of which can be assigned to the most appropriate agent: the model, the clinician, or a hybrid interaction between the two. Prior work such as PACE~\citep{zheng2021pace} demonstrated the value of learnable task decomposition in human-in-the-loop healthcare delivery. Embedding a similar principle into a foundation-model pipeline provides clinicians with explicit mechanisms for overriding, refining, or extending cohort definitions during deployment.
Operationally, this is implemented as a closed-loop workflow in which the computationally expensive backbone is reused, while only cohort retrieval and aggregation are recomputed. For example, a clinician may modify the cohort definition by imposing an additional constraint such as ``include only patients with stage 3 CKD or worse.'' The system then re-filters the retrieval index according to the revised cohort specification, recomputes the within-cohort outcome distribution, and records the modification for subsequent retraining of $g_\phi$. Because the backbone encoder $f_\theta$ does not need to be re-evaluated, the updated cohort summary can be returned at approximately the cost of a database query rather than a full model forward pass.
Workflows that combine human-curated cohort anchors with learned ones~\citep{halpern2016electronic} integrate naturally within this stage and provide a practical entry point for institutions that lack labelled data to train a learned cohort module from scratch.

\mypara{Lifecycle monitoring.}
Foundation models drift~\citep{finlayson2021clinician, guo2022evaluation};
the question is not whether but when. Cohort anchoring provides a natural
drift signal: if the distribution of retrieved nearest cohorts for incoming patients
deviates from the training distribution, or if the within-cohort outcome rates diverge from their training-time estimates, the model flags itself for retraining.
We expect cohort-resolved
drift monitoring to be more sensitive than conventional score-level monitoring in practice, because it
decomposes the patient population along clinically meaningful axes rather than averaging over
heterogeneous subgroups. For example, a four-percentage-point performance change confined to the elderly-with-multiple-comorbidities cohort may be clearly observable at the cohort
level while being partially obscured in aggregate population-level statistics. 
We emphasise that this remains a testable prediction of the framework rather than an established empirical result (Hypothesis~H3, Section~\ref{sec:hypotheses}).
This perspective on continuous monitoring is conceptually aligned with collaborative pipeline-management systems such as MLCask~\citep{luo2021mlcask} and with broader efforts toward reproducible lifecycle management for clinical machine learning systems~\citep{sendak2020realworld, wiens2019no}.

\subsection{Implementation sketch}
\label{sec:framework:algorithm}

Algorithm~\ref{alg:cafm} summarises the overall pretraining and inference procedure. The bespoke component is the \textsc{NearMissNeg} subroutine, formally defined in Section~\ref{sec:framework:pretraining}, which transforms an otherwise generic contrastive objective into a cohort-discriminative learning mechanism. Beyond this component, the overall pipeline largely follows the standard contrastive training paradigm, augmented with cohort-aware positive and negative pools and a cohort-level retrieval index assembled from the representations learned during pretraining.

\begin{algorithm*}[t]
\caption{\cafm: pretraining and inference procedures.}
\label{alg:cafm}
\begin{small}
\mypara{Inputs:} curated EHR corpus $\mathcal{D}$ (Stage 1, with deviation
flags), event-level objective $\mathcal{L}_{\text{seq}}$, modality encoders
$\{f_\theta^{(m)}\}_{m\in M}$, cohort module $g_\phi$, weights $\lambda_p,
\lambda_n, \lambda_b, \{\lambda_m\}$, temperatures $\tau_+, \tau_-$,
near-miss tradeoff $\beta$, momentum $\eta$ for cohort statistics.

\mypara{Warm-start (once before pretraining):} initialise $g_\phi$ to fit a
weakly supervised phenotyping signal (rule-based, anchor-and-learn, or
ontology-derived); compute initial phenotype-overlap matrix $\rho(c, c')$
offline; initialise outcome-proxy estimates $Y(c)$ from corpus statistics.

\mypara{Pretraining:} for each minibatch $\mathcal{B} \subset \mathcal{D}$,
\begin{enumerate}[leftmargin=1.5em,topsep=0pt,itemsep=1pt]
\item Drop deviation-flagged records that fail a clinical-validity rule.
\item Compute soft cohort assignments $\pi_i = g_\phi(x_i)$; form pair
similarities $s_\pi(x_i, x_j)$ and their normalisation
$\tilde{s}_\pi(x_i, x_j)$ over $\mathcal{B}\setminus\{x_i\}$.
\item For each modality $m$, compute embeddings $f_\theta^{(m)}(x_i)$; fuse
and project to the shared embedding $z_i$.
\item Score candidate cohorts $c'$ by $\rho(c_i, c') - \beta\,
D_{\mathrm{KL}}(Y(c_i) \,\|\, Y(c'))$ using the running estimates of $\rho$ and
$Y$; sample $\textsc{NearMissNeg}(x_i)$ above a quantile threshold.
\item Form $\mathcal{L}_{\textsc{cafm}}$ as in Eq.~\ref{eq:cafm-full}
(comprising $\mathcal{L}_{\text{seq}}$, the cohort attraction/repulsion
terms $\bar{\mathcal{L}}_{\text{coh}\pm}$, the cluster-balance term
$\mathcal{L}_{\text{bal}}$, and the per-modality intra-cohort terms
$\sum_m \lambda_m\,\bar{\mathcal{L}}_{\text{intra}}^{(m)}$); update
$(\theta, \phi, \psi)$ with one optimiser step.
\item Update $Y(c) \leftarrow \eta\,Y(c) + (1-\eta)\,\hat{Y}(c \mid
\mathcal{B})$ for each $c$ represented in the batch; periodically refresh
$\rho$.
\item Anneal $\lambda_p, \lambda_n$ along a fixed schedule.
\end{enumerate}

\mypara{Index construction (after pretraining):} compute and store
$(z_i, \pi_i, y_i)$ for all $x_i \in \mathcal{D}$; build an approximate
nearest-neighbour index keyed by $z_i$ and a cohort-level summary table
keyed by $c$ (size, outcome statistics, anchor-feature distributions).

\mypara{Inference for a query patient $x_q$:}
\begin{enumerate}[leftmargin=1.5em,topsep=0pt,itemsep=1pt]
\item Compute $z_q$ and $\pi_q$.
\item Retrieve top-$k$ nearest patients via the index; aggregate by their
hard cohort assignment to identify the cohort(s) supporting the query.
\item Aggregate the within-cohort outcome distribution; report cohort size,
per-cohort calibration, and outcome rate. If the retrieved cohort size falls
below a coverage threshold, flag the prediction for clinician deferral.
\item Optionally, pass the cohort summary to a language-model
summariser~\citep{singhal2025medpalm2, saab2024medgemini} for
clinician-facing narrative.
\item Log cohort statistics for ongoing drift monitoring during deployment.
\end{enumerate}
\end{small}
\end{algorithm*}

The sketch makes three points explicit. First, \cafm operates as a
\emph{retrofit} for existing EHR backbones: one can directly adopt BEHRT~\citep{li2020behrt}, Med-BERT~\citep{rasmy2021medbert}, or MOTOR~\citep{steinberg2024motor} as $f_\theta$ and gain cohort-level functionalities without rebuilding the encoder. 
This retrofit assumes only that the
backbone exposes a fixed-dimensional patient representation (optionally through an additional
pooling layer when the model produces event-level embeddings) and
that its tokenisation scheme remains compatible with the cohort module's input
interface.
Second, the \textsc{NearMissNeg} sampler 
constitutes the operational core of cohort discrimination.
Without this mechanism, the contrastive objective degenerates into standard
instance discrimination~\citep{chen2020simclr}, encouraging the model merely to separate patients from one another rather than to learn clinically meaningful cohort boundaries.
Third, the inference procedure treats cohort-level statistics as the first-class output, with scalar risk scores relegated to derived summaries. This reframing directly supports the auditability, transferability, and monitoring capabilities required for reliable clinical deployment and drift-aware model management.

\section{Open Challenges}
\label{sec:challenges}

Five open challenges will determine whether cohort anchoring can mature into a practically deployable paradigm. For each challenge, we discuss the underlying mechanism that makes the problem difficult, the partial solutions that existing literature already provides, and the specific unresolved gap exposed by the \cafm framework.

\subsection{Data deviation, missingness, and bias}

Large-scale pretraining implicitly assumes that the underlying corpus constitutes, on average, a faithful representation of the target population. EHR corpora violate this assumption in multiple well-documented ways. Demographic and socioeconomic factors influence what clinical information is recorded, when it is recorded, and the degree to which it is recorded accurately~\citep{obermeyer2019dissecting, chen2021ethical,
gianfrancesco2018potential}. 
Pretraining objectives that ignore these biases risk encoding documentation artefacts as if they were genuine clinical signal. Recent analyses~\citep{wornow2023shaky} further demonstrate how often the resulting models are evaluated on tasks that obscure rather
than expose this failure mode. 
Addressing this issue is substantially more difficult than it may initially appear because bias enters the data-generation pipeline at multiple stages, including data collection, clinical coding, care delivery, ordering behaviour, and billing practices. Moreover, the effects introduced at each stage are often partially confounded with one another, making individual sources of bias difficult to isolate or correct.

Three specific gaps remain. First, the deviation-detection mechanism introduced in Stage~1~\citep{zheng2025detecting} addresses only one aspect of the problem: records whose distribution is statistically anomalous, but does not directly address records whose distribution appears consistent with a biased process. A more complete solution therefore requires augmenting deviation detection with mechanism-aware filtering strategies capable of distinguishing ``rare but clinically valid'' observations from ``common but biased'' ones.
Second, tighter integration with data-quality and standardisation frameworks~\citep{kahn2016harmonized, hripcsak2015observational} is necessary to enable reproducible construction of pretraining curricula across institutions. In the absence of standardised curation protocols, inter-institutional comparisons remain fragile and difficult to interpret.
Third, fairness-aware
training objectives~\citep{martinez2020minimax, zhang2022fairCXR} have thus far
been studied primarily at the downstream task level, with relatively limited attention paid to the pretraining-corpus level. 
In particular, we believe there is substantial value in protocols that \emph{declaratively specify} what constitutes a clinically meaningful deviation versus an artefact of data collection or documentation, so that clinicians can audit the curation rules before any pretraining begins. Ad-hoc curation procedures, by contrast, are neither reproducible nor institutionally auditable, making them difficult to validate, compare, or govern.

\subsection{Irregular temporality and disease progression}

Temporal structure in EHR data is simultaneously dense (e.g., frequent vital-sign measurements) and sparse (e.g., infrequent oncology follow-up visits). Foundation models trained over tokenised event sequences struggle to handle this without either excessive padding or loss of clinically meaningful event-time information~\citep{shukla2021multitime, tipirneni2022strats}. However, the underlying challenge is more subtle than ``irregular sampling'' alone. Two patients may exhibit highly similar event sets while corresponding to radically different clinical states because of differences in inter-event timing. For example, a creatinine measurement of 2.0 mg/dL on day 3 of admission with a baseline of 1.0 carries different clinical implications from the same value observed on day 14 with a baseline of 1.8.
Several lines of prior work provide partial solutions to this problem, including methods that explicitly model feature-level temporal irregularity~\citep{zheng2017capturing, bai2018interpretable}, dynamic network adaptation mechanisms~\citep{zheng2022dyhealth}, and continuous-time representation architectures~\citep{rubanova2019latent, kidger2020neural}.

A cohort-anchored model further amplifies this challenge because cohorts themselves are
inherently time-indexed objects. 
A patient’s cohort membership at the beginning of admission is not necessarily the same as her cohort membership 48 hours later, and pretraining must accommodate this temporal evolution. 
The solution we expect to be most promising is \emph{trajectory-conditioned cohort modules}, which emit sequences of cohort assignments along a patient timeline rather than a single static cohort label. Under this formulation, a query issued at hour 48 retrieves a cohort composed of patients who resembled the index patient specifically \emph{at hour 48 of their own clinical trajectories}.
Most downstream components, including the contrastive objective, retrieval index, and monitoring pipeline, generalise naturally from patient-level representations to representations defined over $(\text{patient}, \text{time})$ pairs. The primary costs are increased storage requirements and the need for a more carefully controlled sampling strategy, since trajectories originating from the same patient must be handled cautiously to avoid overwhelming the cohort distribution with near-duplicate instances of the anchor trajectory.
If successful, however, this formulation would allow cohort retrieval to depend jointly on clinical state and clinical timing, rather than treating patient similarity as temporally static.

\subsection{Modality contamination and collapse}

Recent multimodal medical models have demonstrated impressive transfer performance~\citep{lu2024conch, zhang2023biomedclip, huang2023plip, moor2023medflamingo}. However, they frequently suffer from modality collapse, in which a single modality, most commonly free-text clinical notes, dominates the joint representation space~\citep{su2025clear, wang2020what, peng2022balanced, hessel2020emap}. The underlying reason is structural. Free-text notes contain orders of magnitude more information per training example than structured event sequences or individual imaging studies, causing gradients to preferentially propagate through the text encoder unless explicit balancing mechanisms are introduced.
The clinical consequences are further amplified by a mismatch between modality availability during training and deployment. For example, discharge summaries available during training are typically written hours or days after the clinical events whose risk a deployed model is expected to predict at the bedside. As a result, a model that learns to depend on such information during training may encounter little or no informative text at inference time. This phenomenon is more accurately characterised as a covariate shift induced by modality-availability mismatch rather than contamination in the strict sense. Nevertheless, the two failure modes reinforce one another: a contaminated multimodal representation makes the deployment-time absence of the dominant modality substantially more harmful than it would be under a balanced pretraining regime.

Cohort-level alignment provides one potential mitigation strategy. Complementary approaches include modality dropout, gradient-balancing losses~\citep{wang2020what, peng2022balanced}, and gradient-conflict resolution via projections~\citep{yu2020pcgrad}, each designed to prevent any individual modality from monopolising the cohort signal. 
These mechanisms operate at different levels of the learning pipeline: modality dropout reduces over-reliance at the input level, gradient balancing and gradient-conflict resolution act at the optimisation level.
One particularly important open question is whether these mechanisms combine constructively or merely act as partial substitutes for one another.
A second unresolved issue is whether the introduction of cohort-level alignment, a cohort prior
that explicitly preserves intra-modal similarity, may render some of these mechanisms
less necessary and others more important, yet the empirical behaviour of these combinations remains largely unexplored.
Finally,~\citep{su2025clear} provides a complementary perspective worth incorporating into the framework: disagreement across modalities, when properly modelled, may itself constitute an informative clinical signal rather than merely noise.

\subsection{Distribution shift and external validation}

EHR foundation models routinely exhibit substantial performance degradation when ported across institutions~\citep{finlayson2021clinician, futoma2020myth, yang2024demographic}. 
This phenomenon is sufficiently consistent that some studies now regard within-site validation as inadequate in the absence of external institutional evaluation. 
Cohort anchoring offers two complementary advantages in addressing this challenge.
The first is improved drift detection. Deployment-time drifts can be monitored at the cohort level rather than only at the aggregate population level, yielding greater sensitivity because cohort-resolved drifts are not diluted across heterogeneous patient subgroups.
The second is targeted adaptation. Instead of fine-tuning the model on all target-site data, which is computationally expensive and may overfit to institution-specific artefacts, adaptation can be restricted to cohorts that overlap with source-site coverage. Cohorts that are rare or absent in the source institution can then be explicitly flagged for additional curation or validation. 
This strategy aligns with prior work on reweighting and domain adaptation~\citep{ganin2016domain, sagawa2020distributionally, koh2021wilds, zhu2021robustcrossdomain}, while introducing a clinically interpretable adaptation unit to otherwise opaque adaptation losses, for example: ``patients resembling X are common at site B but rare at site A.''

External validation remains the gold standard for evaluating clinical prediction systems~\citep{collins2015tripod, collins2024tripod, gallifant2025tripodllm}, and cohort anchoring should not be viewed as a substitute for it. A model that performs well on cohorts represented at the training institution may still fail on clinically distinct cohorts unique to a new deployment site. What cohort anchoring can reduce, however, is the cost of identifying \emph{which} cohorts at the target institution require additional scrutiny. This capability, in turn, can substantially improve the design efficiency of external-validation studies.
The key unresolved question is quantitative: how much overlap between source-site and target-site cohort distributions is necessary before cohort-targeted fine-tuning becomes more effective than training a new model from scratch, and how does this threshold vary across modalities and clinical tasks? To our knowledge, this question has not yet been systematically studied, despite its likely importance for the practical deployment economics of large-scale clinical foundation models.

\subsection{Evaluation beyond accuracy}

Standard EHR benchmarks~\citep{wornow2023ehrshot, johnson2016mimic,
johnson2023mimiciv, pollard2018eicu} primarily evaluate discrimination performance over a fixed set of downstream tasks.
Under this paradigm, models are rewarded for achieving strong aggregate metrics such as the Area Under the Receiver Operating Characteristic (AUROC) on the average patient population, while many clinically relevant properties remain unmeasured.
Cohort-anchored systems require a substantially richer evaluation framework, and we explicitly outline the dimensions that should be assessed below.

\begin{itemize}[leftmargin=1.5em,topsep=2pt,itemsep=4pt]
\item \textbf{Cohort coverage.} Does every test patient retrieve a cohort of sufficient size and clinical coherence, or can the system appropriately defer when no comparable peer population exists? 
Coverage failure represents one of the most common silent failure modes in similarity-based prediction systems: rare or atypical patients may retrieve small and heterogeneous cohorts, leading the model to produce confidently miscalibrated predictions despite weak evidentiary support.

\item \textbf{Cohort calibration.} 
Does the observed outcome rate within each cohort align with the model's predicted probability across cohorts with varying sizes and degrees of homogeneity? Population-level calibration plots may obscure systematic miscalibration within specific cohorts, whereas cohort-level calibration analysis makes such failures explicit.

\item \textbf{Cohort stability.} Does a small perturbation in the input move a patient to a clinically meaningful neighbouring cohort, or merely to a noisy neighbour? This evaluation is inherently local: for example, a one-day shift in admission timing should not cause the retrieved cohort to cross a clinically meaningful boundary unless that boundary itself depends on admission timing.

\item \textbf{Actionability.} Does cohort-level evidence influence clinician decisions in prospective deployment settings, rather than merely improving retrospective AUROC? 
The ultimate evaluation of cohort anchoring is therefore workflow-level: if clinicians consistently find the cohort summary unhelpful or non-informative, they are revealing a limitation that conventional discrimination metrics cannot capture.

\end{itemize}

Recent calls for clinically grounded
evaluation~\citep{ghassemi2021false, raji2022fallacy,
wornow2023shaky} converge on the same underlying need. Cohort anchoring provides a
particularly natural evaluation unit because each metric described above corresponds directly to a specific stage of the framework.
Prospective deployment studies, although still relatively rare, are becoming increasingly
common~\citep{tomasev2019nature, escobar2020automated, sendak2020realworld} and should ultimately serve as the primary standard for evaluation. 
Reporting
frameworks such as TRIPOD+AI~\citep{collins2024tripod} and
TRIPOD-LLM~\citep{gallifant2025tripodllm} establish a minimum standard for what constitutes publishable evidence in this setting.
We further anticipate
that the next generation of EHR benchmarks will incorporate cohort coverage
and cohort calibration as mandatory auxiliary metrics, on the grounds that calibration
claims cannot be meaningfully assessed when the coverage properties of the
retrieved cohorts remain unknown.

\section{Case Studies}
\label{sec:case_studies}

\begin{figure*}[t]
\centering
\includegraphics[width=\textwidth]{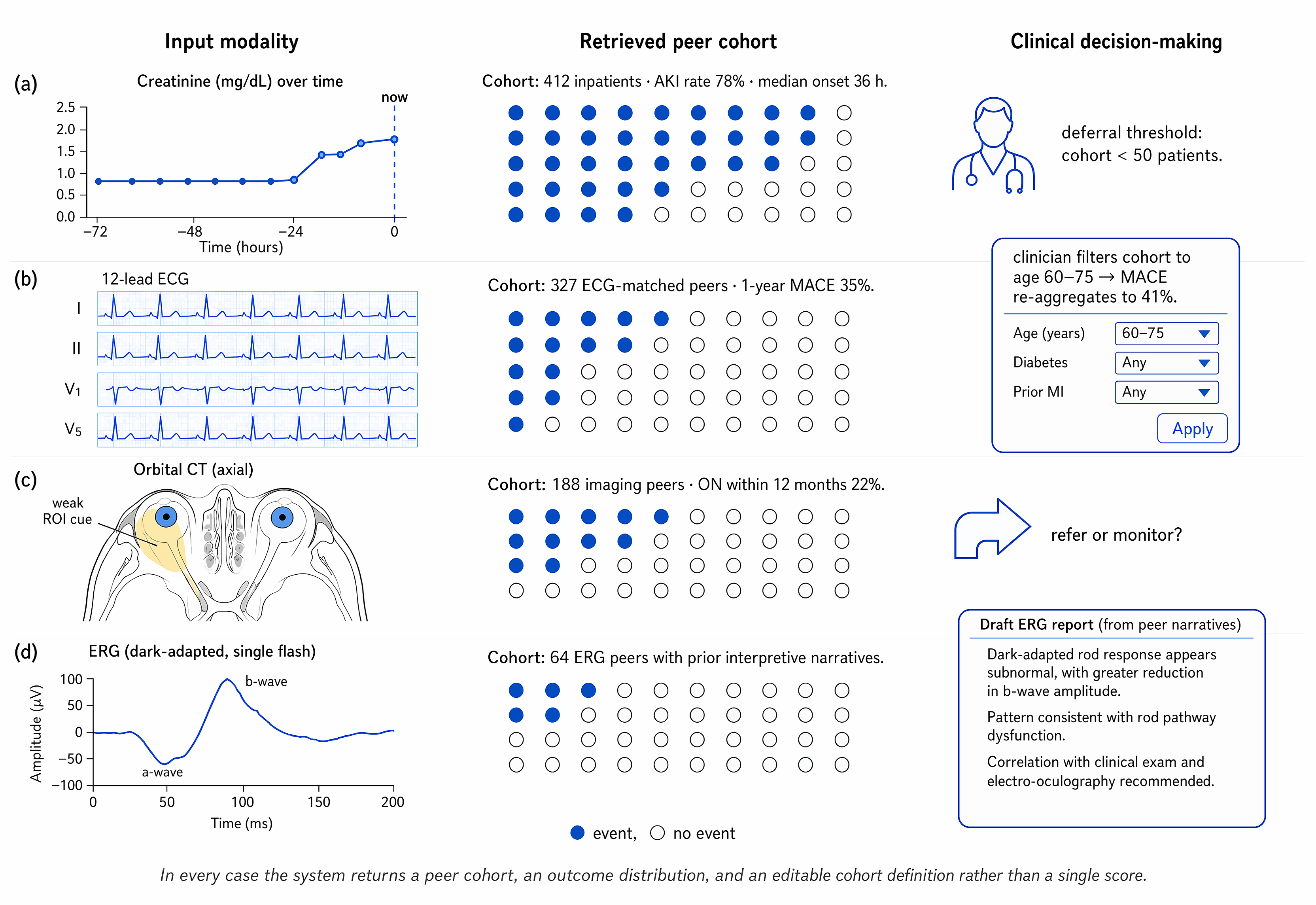}
\caption{The four case studies are illustrated, spanning the
principal modalities of EHR data: 
(a) acute kidney injury prediction
from longitudinal laboratory measurements; 
(b) cardiovascular risk
stratification from electrocardiograms fused with structured records; 
(c) optic neuropathy triage from orbital imaging. In each case, the model returns a peer cohort, an
outcome distribution, and an editable cohort definition rather than a single
score;
and (d) electroretinogram-grounded report generation
extends the same logic to multimodal narrative output.}
\label{fig:case_studies}
\end{figure*}

We illustrate the proposed framework through four case studies spanning the principal modalities encountered in EHR-based clinical AI: structured longitudinal laboratory data, physiological signals, medical imaging, and multimodal report generation (Figure~\ref{fig:case_studies}). Each case study is grounded in operational projects with which we have direct experience and is selected to highlight a distinct capability enabled by cohort anchoring: calibration support in AKI prediction, multimodal triage support in cardiovascular risk assessment, cross-domain transfer robustness in optic neuropathy imaging, and generative grounding in ERG report generation.
The case studies also vary in deployment maturity, ranging from operational systems to active research prototypes. In each setting, we explicitly discuss both the advantages introduced by cohort anchoring and the associated computational, operational, or modelling costs, allowing readers to weigh these benefits against their associated costs.

\subsection{Acute kidney injury prediction}

AKI (acute kidney injury) is a major contributor to in-hospital mortality, despite being largely preventable with timely
intervention~\citep{kellum2021kdigo}. The clinical rationale for early prediction is therefore clear: nephrotoxic medications can be discontinued, fluid management can be adjusted, and renal-replacement therapy can be prepared, but only if risk estimates are generated sufficiently early to support intervention.
\citet{tomasev2019nature} demonstrated the feasibility of clinically relevant continuous AKI prediction using data from the United States Department of Veterans Affairs, achieving 48-hour-ahead prediction for the majority of inpatient AKI episodes. We subsequently adapted this line of work to a Singaporean hospital setting with explicit modelling of both time-invariant and time-varying feature importance~\citep{chua2021healthcare}. This experience highlighted not only the clinical potential of early AKI prediction, but also the limitations of conventional single-score frameworks.

The cohort-anchored perspective changes the capability of the prediction more than its raw predictive accuracy. Rather than returning a single 48-hour AKI risk score, the system retrieves a cohort of prior inpatients with similar admission characteristics and temporal trajectories, together with the within-cohort AKI incidence, median time-to-onset, and the distribution of clinically relevant contributing factors among cohort members who subsequently developed AKI.
This representation naturally supports the types of follow-up questions clinicians routinely ask in practice: ``Were comparable patients receiving nephrotoxic medications?'', ``Which contrast agents were administered?'', or ``How did similar patients respond to fluid challenge?'' Instead of requiring a separate analytics workflow, the cohort representation itself provides the evidentiary substrate needed to answer these questions through direct interrogation of the retrieved cohort.
In effect, the prediction ceases to function as an opaque scalar output and instead becomes a compact, structured clinical evidence object that clinicians can inspect, query, and act upon at the point of care.

This perspective also reveals failure modes that would remain hidden under conventional single-score prediction. Cohorts that are either too small or excessively heterogeneous are likely to exhibit poor calibration: small cohorts produce high-variance outcome estimates, while heterogeneous cohorts aggregate patients whose clinical trajectories diverge sufficiently that the resulting average becomes clinically uninformative.
An immediate operational implication is the introduction of a cohort-coverage threshold below which the system should abstain from prediction rather than return a potentially unreliable estimate. This aligns naturally with the second-opinion paradigm proposed by \citet{kompa2021second}, while grounding abstention in a clinically interpretable unit that clinicians can directly inspect. Whether this relationship holds quantitatively is one of the framework's central empirical hypotheses; Section~\ref{sec:hypotheses} (H1) describes the corresponding details.
The primary cost of cohort anchoring in this setting lies in the additional infrastructure required to expose cohort-level evidence to clinicians. This includes richer visualisation interfaces, cohort-oriented dashboards, and workflow training for staff who must interpret structured cohort summaries rather than single scalar risk values. In ongoing work, researchers are further exploring alignment between cohort retrieval mechanisms and established phenotyping ontologies~\citep{denny2010phewas, choi2017gram}.

\subsection{Cardiovascular risk from electrocardiograms}

ECGs (electrocardiograms) are inexpensive, widely available, and contain substantial physiological signal that often extends beyond what can be reliably interpreted by the unaided human reader. As a result, the clinical case for ECG-based AI is particularly compelling: the marginal acquisition cost of an ECG is effectively negligible, the informational value available to deep learning systems is high, and the downstream clinical interventions triggered by high-risk findings are already well established in routine care.
Deep learning models have achieved cardiologist-level or superior performance on rhythm classification tasks~\citep{hannun2019cardiologist, ribeiro2020automatic, strodthoff2021ptbxl}, while also demonstrating the ability to identify latent conditions such as left ventricular dysfunction~\citep{attia2019artificial} and silent atrial fibrillation~\citep{attia2019afib}, which are difficult for human experts to infer directly from standard 12-lead tracings. More recently, ECG-specific foundation models~\citep{mckeen2024ecgfm, na2024stmem} have extended large-scale self-supervised learning paradigms to this modality, further increasing the potential performance ceiling for ECG-based clinical prediction systems.

Cohort anchoring introduces two important capabilities in this domain that are not captured by a purely single-modality perspective. The first is multimodal alignment. An ECG-derived cardiovascular risk estimate is clinically informative on its own, but becomes substantially more valuable when integrated with structured EHR information such as demographics, comorbidities, medications, and longitudinal laboratory trends that contextualise the ECG within the patient's broader clinical state.
Such fusion is technically non-trivial. The high-density ECG signal can dominate the joint representation during optimisation, overwhelming structured modalities through disproportionately strong gradient signals~\citep{su2025clear, wang2020what}. Building on prior work on adaptive interaction modelling in healthcare	data~\citep{cai2022elda, cai2021armnet}, the cohort-level contamination-control mechanisms introduced in Stage~3 (Section~\ref{sec:framework:multimodal}) provide a natural extension by treating the cohort as the explicit unit of intra-modal structure preservation.
The second capability is clinically grounded triage retrieval. Instead of producing a scalar prediction such as ``one-year MACE risk = 0.35,'' the system retrieves a peer cohort whose ECG patterns resemble those of the index patient \emph{and} whose downstream clinical trajectories are documented within the same institution. The returned evidence therefore includes not only the frequency of major adverse cardiovascular events within the cohort, but also their temporal profiles and treatment responses. In this way, the prediction is grounded in institution-specific historical experience rather than a global population prior~\citep{liu2025neuralcohort, cai2024cohortnet}, organising the ensuing clinical discussion around comparable patient trajectories and treatment outcomes rather than an isolated scalar risk estimate.

The cost of this reframing is non-trivial. Cohort retrieval over combined ECG and EHR representations requires a multimodal retrieval index that is substantially more complex and resource-intensive to construct and maintain than a single-modality alternative. In addition, institutions must commit to maintaining cohort definitions as evolving clinical objects rather than treating them as static artefacts generated once during training.
The anticipated benefit, however, is that this reframing makes deployment-time interactions with cardiologists qualitatively more natural and clinically interpretable, because cohorts constitute a form of evidence that clinicians already routinely reason about in practice. Whether this improvement in interpretability and communication ultimately translates into greater clinical actionability is precisely the question that hypothesis~H5 (Section~\ref{sec:hypotheses}) is designed to test.

\subsection{Optic neuropathy triage from orbital imaging}

Thyroid eye disease can progress to dysthyroid optic neuropathy (DON), a sight-threatening complication for which timely diagnosis is often limited by access to subspecialty expertise~\citep{chng2023application, kheok2025orbitalct}. In practice, the primary bottleneck is not the absence of capable algorithms, but rather the logistics of referral and triage. In many healthcare systems, patients at greatest risk are geographically distant from tertiary or quaternary ophthalmology centres, making the clinically relevant question not simply whether DON is present, but which patients require urgent referral versus continued local monitoring.
Orbital CT-based deep learning models have already demonstrated specialist-level performance for optic neuropathy prediction in quaternary referral cohorts~\citep{kheok2025orbitalct}, while broader advances in medical vision-language pretraining have further reduced the data limitations traditionally associated with ophthalmic AI~\citep{zhang2023biomedclip, zhou2023retfound}. The remaining bottleneck is therefore fundamentally one of clinical trust: can a primary-care clinician or general ophthalmologist confidently act on an algorithmic triage recommendation when the consequence of a false negative may be irreversible vision loss in the affected eye?

The cohort-anchored framing is particularly relevant in this setting because triage is fundamentally a referral problem rather than a pure prediction problem. A primary-care clinician does not merely need a probabilistic estimate; instead, the clinically useful question is: ``Among prior patients resembling this index case, how many were ultimately diagnosed with DON, and what imaging characteristics were observed in those cases?''
Multimodal cohort retrieval over imaging and structured clinical records, combined with weakly supervised segmentation cues for region-of-interest extraction~\citep{su2026discern}, naturally supports this workflow. At inference time, the returned cohort summary effectively functions as a curated set of prior referral cases together with their downstream outcomes, presented at a level of abstraction that clinicians can meaningfully inspect and interrogate. In this sense, the model does not merely issue a prediction; rather, it grounds and delegates its recommendation to the cohort evidence supporting it.

The primary cost of cohort anchoring in imaging applications arises from the scale of the retrieval index itself. Orbital CT studies are storage-intensive, and institutions with relatively modest patient volumes may not possess enough labelled positive cases to construct clinically meaningful cohorts. As a result, practical deployment of cohort-anchored imaging systems may ultimately require federation of cohort indices across institutions to reach the cohort sizes
clinical actionability requires. While such a federation is technically feasible at the embedding level, it becomes substantially more complex when patient-level data sharing is involved because of privacy, governance, and regulatory constraints~\citep{gerke2020ethical}.
This consideration extends naturally beyond orbital imaging to related ophthalmic applications addressed by retinal foundation models~\citep{zhou2023retfound}, and motivates the transition to the fourth case study.

\subsection{Electroretinogram-grounded reports}
\label{sec:case_studies:erg}

ERGs (electroretinograms) are electrophysiological assessments of retinal and visual-pathway function that play a central role in diagnosing and managing a range of ophthalmic disorders. Despite their clinical importance, ERGs are rarely analysed at scale because interpretation is highly specialised and time-intensive. In practice, ophthalmologists interpreting an ERG consider not only the waveform morphology itself, but also the patient's clinical history, relevant imaging findings, and prior experience with similar cases.
The resulting report is therefore inherently synthetic: it identifies the abnormality, contextualises the findings within the patient's broader clinical presentation, and recommends an appropriate clinical course of action. Producing clinically usable ERG reports at the model level consequently requires substantially more than a high-capacity medical language model~\citep{singhal2025medpalm2, saab2024medgemini, lievin2024canllm}. The generated report must instead be \emph{grounded} simultaneously in measurable electrophysiological signal characteristics, correlated imaging evidence~\citep{zhou2023retfound}, and the documented trajectories of comparable historical cases.

Cohort anchoring provides the missing grounding mechanism. At inference time, the cohort retrieval index returns prior ERG cases that most closely resemble the index examination, together with their associated interpretive narratives written by ophthalmologists during previous consultations. These retrieved cases then form a curated grounding corpus over which a language-model summariser generates the final report \emph{conditioned on} clinically comparable patient trajectories rather than relying solely on parametric memory. Conceptually, this follows the broader paradigm of retrieval-augmented generation~\citep{lewis2020rag, gao2024retrievalsurvey, xiong2024medrag, xu2024ramehr}, 
but replaces text passages with patients as the retrieval unit.
This distinction is structurally important. A document-grounded summariser retrieves passages that are linguistically similar to the query, whereas cohort-grounded report generation retrieves \emph{patients} whose underlying clinical trajectories resemble those of the index case, and then uses the associated historical reports as the evidentiary basis for generation. The difference becomes particularly apparent in the resulting failure modes. A document-grounded model may hallucinate by combining linguistically related yet clinically incompatible passages, whereas a cohort-grounded model is constrained to synthesise narratives from comparable patient populations.

The resulting report is accompanied by the cohort evidence from which it was generated, making the generation process auditable in precisely the manner envisioned in Stage~4 (Section~\ref{sec:framework:in-the-loop}). A clinician reviewing the report can inspect the supporting peer cases, assess whether they are clinically relevant, and then either accept the generated interpretation or escalate the case for further review.
Among the four case studies, this scenario is the most forward-looking, and we therefore present it primarily as a proof of concept illustrating how cohort anchoring may reshape generative clinical AI in the same way that it reshapes discriminative prediction systems. At present, the supporting evidence remains preliminary. The central open question is whether the narrative quality and clinical reliability of cohort-grounded report generation can be maintained at the scale and variability of routine clinical deployment.
Our working hypothesis is cautiously optimistic, based on the intuition that cohort grounding imposes on a language model a constraint analogous to the case-based reasoning structure that an experienced ophthalmologist provides to a junior one. Whether this intuition remains valid under prospective evaluation is the empirical question we aim to investigate next.

\section{What Cohort Anchoring Is and Is Not}
\label{sec:discussion}

Cohort anchoring sits adjacent to several active lines of work and a perspective
paper 
must clearly delineate its scope to avoid the impression that it merely repackages ideas already familiar to the community.
We therefore devote this section to the boundaries themselves, examining each adjacent technique as a potential alternative to cohort anchoring and explaining why we view it as complementary rather than substitutive. 
The intended audience is the experienced reader inclined to ask a natural question: ``Why is this problem not already solved?''

\mypara{Not retrieval-augmented generation.}
Retrieval-augmented generation~\citep{lewis2020rag, gao2024retrievalsurvey,
xiong2024medrag, xu2024ramehr} retrieves \emph{documents} or knowledge snippets
to ground the output of a language model. The retrieved unit is textual, such as a clinical guideline passage, textbook excerpt, or section of a discharge summary, and is incorporated into the model's context during generation. 
Cohort anchoring, by contrast, retrieves \emph{patient cohorts}. 
These retrieved objects are structured rather than textual, summarise clinically relevant outcome distributions rather than passages, and remain auditable through their underlying diagnoses, laboratory measurements, treatments, and outcomes.
The distinction is substantive rather than cosmetic. 
A document cannot be partitioned by outcome; a cohort can. 
A document cannot answer ``how many of those patients also had stage 3 CKD?''; a cohort can. 
A document cannot be edited by a clinician mid-consultation in a way that changes the model's prediction; a cohort can.
The two techniques are complementary: cohort retrieval can itself serve as the grounding mechanism for a capable medical language model~\citep{singhal2025medpalm2, saab2024medgemini}, which converts the retrieved structured cohort evidence into a clinician-readable narrative, as illustrated in the ERG case study (Section~\ref{sec:case_studies:erg}).

\mypara{Not patient-level kNN over learned embeddings.}
The closest technical analogue to cohort anchoring is patient-level retrieval
using nearest neighbours in a learned embedding
space~\citep{suo2018deep, landi2020deep}. Although the two approaches share retrieval
infrastructure, they differ fundamentally in both what is retrieved and what is reported to the user.
Patient-level kNN retrieval returns a set of $k$ nearby patient representations whose embeddings lie close to the query in latent space. The retrieved set is not labelled, not addressable, and not editable. Cohort anchoring, in contrast, retrieves a discrete cohort identifier and then accesses the corresponding cohort membership and within-cohort outcome distribution.
The retrieved cohort is explicitly defined through a construction mechanism such as a rule set, prototype, or ontology-based grouping, 
making it inspectable (every member is recoverable from the index) and editable (a clinician can remove members or refine the cohort definition before finalising a decision). 
The distinction is therefore the difference between ``these twenty patients are nearby in
embedding space'' and ``these patients form a cohort characterised as
\emph{community-acquired pneumonia in the elderly}, within which 78\% experienced
the outcome of interest.'' The former is a debugging tool; the latter is clinical evidence. 
We expect both to coexist in practical systems. Patient-level kNN retrieval remains valuable for representation analysis, debugging, and validation of the cohort construction process, whereas cohort anchoring provides the structured, population-level evidence required for clinician-facing decision support.

\mypara{Not in-context learning.}
Few-shot prompting with patient examples~\citep{brown2020gpt3,
glicksberg2024evaluating} can mimic cohort-based reasoning within the context window of a large language model. 
By providing several clinically similar patient summaries together with their outcomes and asking the model to reason about an index patient, one can achieve behaviour that superficially resembles cohort retrieval. 
The appeal of this approach is obvious: it requires neither model retraining nor dedicated retrieval infrastructure.
However, we believe this is structurally limited for three reasons. 
First, context windows remain finite. Even large-context models can accommodate only a modest number of patient examples, far fewer than the cohort sizes typically required to support reliable calibration and outcome estimation. 
Second, in-context learning lacks an explicit mechanism for modelling near-miss examples. The prompt is presented as a flat collection of demonstrations, and the model is not naturally informed that certain examples are intentionally included as clinically similar yet outcome-divergent contrasts.
Third, in-context learning offers no inherent support for lifecycle monitoring. When deployment-time distributions shift, the system produces no corresponding drift signal unless the prompts themselves are revised, and there is no principled mechanism through which the system can indicate that retraining or recalibration is required.
By contrast, cohort anchoring offers a structural commitment that survives context resets and
remains addressable for the auditing and drift-monitoring capabilities described in
Stage~4 (Section~\ref{sec:framework:in-the-loop}). 
The two approaches are complementary rather than competing. A cohort-anchored system may still employ in-context learning within its language-model layer, but the underlying commitment to cohort structure must reside at the system level rather than the prompt level.

\mypara{Not post-hoc explanation.}
Saliency maps, attention visualisations, and Shapley-style attribution methods seek to explain individual predictions by identifying the input features that most strongly influenced the model's output~\citep{lundberg2017shap, ribeiro2016should}.
The motivation is well-founded: to provide users with some understanding of why a prediction was made.
The underlying mechanism, however, is a retrofit. 
A model trained without an intrinsic notion of evidence is subsequently asked, after the fact, to reconstruct one, and the resulting explanation inevitably inherits the limitations of the learned
representation.
Existing studies have documented the failure modes: 
post-hoc explanations can be invariant to model parameters~\citep{adebayo2018sanity}, 
sensitive to clinically irrelevant perturbations~\citep{ghorbani2019interpretation}, 
and only weakly aligned with clinically meaningful features~\citep{saporta2022benchmarking}. 
For a clinical perspective on these limitations and their implications for healthcare applications, we refer readers to~\citep{ghassemi2021false}.
More broadly, \citet{rudin2019stop} argues that the field should prioritise models that are intrinsically interpretable by design rather than relying on post-hoc explanatory mechanisms to justify opaque predictions.
Cohort anchoring represents one concrete realisation of this principle.
Rather than attempting to reconstruct an explanation after the prediction has been produced, the framework grounds the prediction itself in an explicitly retrievable patient population. Under this view, explanation is not an additional artefact generated after inference; the cohort itself \emph{is} the explanation.
This does not render feature-level attribution obsolete. Attribution methods remain valuable for tasks such as debugging input pipelines, identifying spurious correlations, and understanding representation behaviour. However, the strategic interpretability commitment of \cafm resides at the cohort level rather than the feature level: the primary explanatory unit is a clinically meaningful population, not a ranked list of influential input variables.

\mypara{Not prototype networks alone.}
Prototype-based interpretable
models~\citep{ming2020protosteer, chen2019looks} are
an important conceptual precursor to cohort anchoring. 
Both approaches share a commitment to representing the training distribution through discrete, inspectable entities rather than relying exclusively on opaque latent representations. 
However, cohort anchoring extends prototype-based reasoning in several ways that become particularly important at the scale of foundation-model pretraining and deployment.
First, cohort membership is conditioned on
\emph{deviation-aware curation} (Section~\ref{sec:framework:stage1}).
At the corpus scale, manual inspection is infeasible, and therefore the quality of the retrieved cohorts depends critically on the quality of the curation process that defines them. Cohort construction must consequently inherit the same data-quality and bias-mitigation principles applied during corpus preparation.
Second, the contrastive head treats near-miss cohorts as first-class negatives~\citep{zheng2024exploiting}.
This recovers a highly informative region of the contrastive space that prototype-based approaches have traditionally collapsed into a generic ``everything else'' category.
Third, the multimodal alignment in Stage~3 (Section~\ref{sec:framework:multimodal}) makes
prototypes coherent across modalities, so that the same prototype unit can serve retrieval across structured EHR data, medical imaging, and physiological signals.
\cafm's contribution lies in this integration: a cohort module that
inherits the commitments of the patient-similarity literature while adapting them
to the scale and operational requirements of modern clinical foundation models.

\mypara{Not causal inference.}
A natural challenge is that cohort retrieval supplies association, not
causation. If a retrieved cohort summary indicates that ``78\% of similar patients developed
the outcome,'' a clinician must still ask whether this observation reflects an intrinsic property of the cohort as a population or a property of the treatment policies under which
the cohort was managed.
Counterfactual decision-support frameworks~\citep{schulam2017reliable} address precisely this distinction by estimating outcomes under hypothetical treatment policies, and the broader
literature on individualised treatment-effect estimation in medicine has evolved into a substantial research area with dedicated methodologies, benchmarks, and evaluation protocols.
We therefore do not view cohort anchoring as a substitute for causal inference. Rather, we view it as a substrate that makes causal questions easier to formulate and investigate.
A retrieved cohort with a well-documented treatment distribution provides a natural input to a counterfactual outcome estimator. 
Moreover, the auditing capabilities introduced in Stage~4 (Section~\ref{sec:framework:in-the-loop}) expose exactly the population-level information that causal analysts typically require when assessing the validity of an inference. 
A mature deployment would compose cohort retrieval with a counterfactual
head that re-expresses the within-cohort outcome distribution as an effect
conditional on the treatment policy.

\mypara{What it requires.}
Cohort anchoring introduces several practical costs that should be explicitly considered by practitioners evaluating its adoption.
First, it requires dedicated infrastructure. The framework requires the institution to construct and maintain large patient repositories that remain available throughout deployment, building on advances in healthcare data systems and integrative analytics platforms~\citep{lee2017bigdata, jagadish2014bigdata, ooi2015singa, zheng2022dyhealth}. The associated cohort retrieval index incurs costs related to construction, maintenance, governance, and security.
Second, cohort anchoring requires sustained attention to data quality. Deviation detection and data-quality monitoring~\citep{zheng2025detecting, weiskopf2013methods} must be treated as continuous processes, rather than one-time preprocessing steps, because clinical documentation practices, coding behaviours, and care pathways evolve over time.
Third, successful deployment depends on clinician engagement. Cohort definitions must be reviewed, validated, and periodically revised to remain aligned with contemporary clinical practice. This introduces an ongoing demand on clinical expertise and institutional resources rather than a one-time development cost.
These costs differ in character. The infrastructure burden is primarily an engineering challenge, whereas data-quality governance and clinician involvement are fundamentally sociotechnical concerns~\citep{sendak2020realworld, gerke2020ethical}.
The expected return is in deployment-time savings: cohort anchoring lowers
the cost of external validation, clinician onboarding, regulatory
documentation, and post-deployment surveillance, each of which currently
dominates the operational budget of a clinical AI system. 
We expose this trade as a working hypothesis of the framework rather than as an established empirical conclusion.

\mypara{Privacy, consent, and federation.}
A cohort retrieval index is, by construction, an indexable record of historical patient populations and therefore inherits the privacy obligations of any system built upon identifiable clinical data. 
Consequently, deployments must ensure that cohort summaries returned at inference time do not disclose identifiable information beyond the scope permitted by institutional policies and data-sharing agreements.
In addition, retrieval logs themselves must be treated as sensitive assets, as they may introduce new attack surfaces if not appropriately secured.
For multi-institution settings, federated learning~\citep{rieke2020federated} provides a natural implementation pathway: cohort modules can be trained over distributed corpora
without centralising patient-level data, while retrieval can operate over a
distributed index whose primitives are carefully scoped. 
The privacy contribution of cohort anchoring is inspectability: the unit of retrieval is a clinically meaningful cohort that a privacy officer can reason about directly, rather than an opaque embedding cluster legible only to developers. 
Ultimately, the governance requirements for cohort-anchored systems overlap substantially with those already facing large-scale clinical AI deployments~\citep{gerke2020ethical}. As the underlying technical capabilities mature, privacy, consent, and federation frameworks will need to evolve in parallel to ensure that the resulting systems remain both clinically useful and institutionally accountable.

\subsection{Empirical hypotheses}
\label{sec:hypotheses}

The framework predicts five empirical regularities, each formulated as a falsifiable hypothesis that can be evaluated through a reasonably sized experimental study. 
For each hypothesis, we specify the expected effect, outline an appropriate experimental design, and identify the negative result that would constitute a meaningful refutation of the framework's prediction.

\mypara{H1 (cohort coverage as a calibration predictor).}
Define the cohort coverage of a query patient as $|\{x_j \in \mathcal{D} : s_\pi(x_q,
x_j) > \kappa\}|$ for a fixed similarity threshold $\kappa$.
We hypothesise that, on a held-out evaluation set, cohort coverage will be negatively associated with prediction calibration error. Specifically, patients who retrieve small or low-density cohorts are expected to exhibit higher expected calibration error (ECE) than patients supported by larger and more densely populated cohorts.
\emph{Design:} Partition test patients into coverage-based strata, compute ECE within each stratum, and evaluate whether calibration error decreases monotonically as cohort coverage increases.
\emph{Refutation:} The hypothesis is refuted if no statistically significant relationship is observed between cohort coverage and ECE, or if the observed association is consistently in the opposite direction.
\emph{Sensitivity:} The result should remain qualitatively consistent across a pre-registered range of $\kappa$ values rather than depending on a single threshold choice, thereby reducing the possibility of threshold-specific optimisation.

\mypara{H2 (near-miss negatives outperform random negatives).}
Holding the backbone architecture fixed, we hypothesise that replacing randomly sampled negatives with near-miss cohort negatives in Stage~2 (Section~\ref{sec:framework:pretraining}) will improve few-shot transfer performance on underrepresented downstream tasks.
\emph{Prior support:} \citet{zheng2024exploiting} demonstrates that near-miss negatives can sharpen cohort boundaries in the cohort-discovery setting. The present hypothesis extends that intuition to the foundation-model pretraining setting, which has not been directly evaluated in prior work to our knowledge.
\emph{Design:} Pretrain two otherwise identical foundation models that differ only in their negative-sampling strategy. One model uses randomly sampled negatives, while the other employs the proposed near-miss cohort negatives. The resulting representations are then evaluated using few-shot AUROC on underrepresented tasks drawn from the long tail of the EHRSHOT benchmark~\citep{wornow2023ehrshot}.
\emph{Refutation:} The hypothesis is refuted if the near-miss sampling strategy fails to improve few-shot performance, if gains are restricted to high-resource tasks in the head of the task distribution, or if the observed improvements disappear when the cohort module is ablated.

\mypara{H3 (cohort drift detects deployment shift earlier than score drift).}
We hypothesise that the divergence between the deployment-time and training-time cohort distributions will serve as an earlier indicator of distribution shift than conventional performance-based monitoring~\citep{finlayson2021clinician, guo2022evaluation}. Specifically, the KL divergence between the deployment-time distribution of cohort assignments and the corresponding training-time distribution is expected to provide advance warning of clinically meaningful drift before substantial degradation becomes apparent in aggregate performance metrics such as AUROC.
\emph{Design:} Monitor both cohort-distribution drift and conventional score-level performance drift over the same prospective deployment period. For each signal, measure the lead time relative to a clinician-validated drift event. The primary comparison is the extent to which each monitoring strategy provides actionable early warning.
\emph{Refutation:} The hypothesis is refuted if conventional score-level monitoring detects drift earlier than cohort-level monitoring, achieves higher precision in identifying clinically meaningful drift events, or consistently provides equivalent lead times without the additional complexity introduced by cohort-based monitoring.

\mypara{H4 (cohort-level alignment mitigates modality collapse).}
We hypothesise that cohort-level multimodal alignment can reduce modality collapse without sacrificing overall predictive performance. Specifically, adding the per-modality intra-cohort loss
$\sum_m \lambda_m\,\bar{\mathcal{L}}_{\text{intra}}^{(m)}$ 
introduced in Section~\ref{sec:framework:multimodal} (Stage~3) is expected to reduce the performance disparity between the strongest and weakest modalities~\citep{su2025clear, wang2020what}, while preserving overall task performance.
\emph{Design:} Compare otherwise identical multimodal models with and without the intra-cohort term while holding $\mathcal{L}_{\text{seq}}$ and the two cohort terms
$\mathcal{L}_{\text{coh+}}, \mathcal{L}_{\text{coh-}}$ fixed.
Performance is then evaluated both at the joint-model level and at the individual-modality level, using solo-AUROC for each modality as a measure of retained discriminative capability. The primary outcome is the dominance ratio between the strongest and weakest modality.
\emph{Refutation:} The hypothesis is refuted if the dominance ratio remains unchanged or increases after introducing the intra-cohort alignment term, or if the resulting reduction in modality imbalance comes at the cost of a statistically or clinically meaningful degradation in overall joint-model AUROC beyond a pre-registered threshold.

\mypara{H5 (clinicians prefer cohort evidence to scores in deferred-decision
contexts).}
We hypothesise that, in clinical scenarios where uncertainty is sufficiently high to justify consultation, escalation, or deferral, clinicians will prefer cohort-based evidence over standalone risk scores. Specifically, when presented with identical underlying predictions, clinicians who receive a peer cohort together with its associated outcome distribution are expected to report greater decision confidence and to make more appropriate deferral decisions than clinicians who receive only a scalar risk estimate.
\emph{Design:} Conduct a randomised between-subjects user study involving clinicians from specialties such as cardiology, nephrology, and ophthalmology. Participants are presented with identical clinical scenarios and underlying model outputs, differing only in how evidence is communicated. The primary outcomes include decision confidence, decision quality, and the appropriateness of referral, escalation, or deferral decisions.
\emph{Refutation:} The hypothesis is refuted if clinicians exposed to cohort-based evidence show no measurable improvement in confidence or decision quality relative to the risk-score condition, or if cohort-based evidence leads to less appropriate decisions despite providing additional contextual information.

Each hypothesis represents a concrete, testable commitment made by the framework to the community. In principle, any individual hypothesis can be evaluated by a single research group within a few months of focused effort. Moreover, a rigorous refutation would help pinpoint the specific stage of the framework responsible for the observed failure.

\subsection{Scope and caveats}
\label{sec:limitations}

Before \cafm is adopted in practice, four caveats warrant careful consideration. Each highlights a boundary condition under which the framework's guarantees may be weakened or its underlying assumptions require additional empirical validation.

\mypara{Scaling of the cohort count.}
The framework does not prescribe how the number of cohorts, $K = |\mathcal{C}|$, should be chosen. When $K$ is on the order of a few thousand, the cohort-cohort similarity matrix $\rho$ remains computationally tractable. 
However, as $K$ grows substantially larger, the $O(K^2)$ cost of precomputation and storage may become a bottleneck, necessitating approximate methods. Determining the appropriate value of $K$ for a given clinical task remains an open research question. One plausible, yet unexplored, direction is the development of hierarchical cohort modules that operate at multiple levels of granularity simultaneously.

\mypara{The wrapper claim has caveats.}
Section~\ref{sec:framework:algorithm} presents \cafm as a retrofit that can be integrated with an existing backbone. However, this claim relies on several assumptions. Specifically, the backbone must either provide fixed-dimensional patient representations or support an appropriate pooling mechanism, and its tokenisation scheme must be compatible with the cohort module. Architectures that produce only event-level outputs or employ highly specialised tokenisation strategies may require substantial adapter engineering. Quantifying this integration cost is beyond the scope of the current framework.

\mypara{Outcome proxy and the supervision boundary.}
The near-miss sampling mechanism relies on cohort-level outcome statistics, $Y(c)$. When $Y$ is derived from labelled clinical outcomes, the framework moves beyond purely self-supervised learning and enters the regime of multi-task supervised pretraining. While this is a reasonable and defensible design choice, it represents a substantive shift in the learning paradigm and should be made explicit when the framework is evaluated or deployed in real-world clinical settings.

\mypara{Privacy scope.}
Although cohort retrieval is more transparent and inspectable than opaque embedding-based retrieval, it does not alter the underlying privacy requirements. Cohort summaries are ultimately derived from identifiable patient records and therefore remain subject to the same institutional, regulatory, and legal safeguards that govern retrieval-augmented clinical AI systems more broadly. The contribution of the framework in this regard is improved interpretive inspectability of the retrieval unit itself, rather than the introduction of new governance or compliance mechanisms.

\section{Related Work}
\label{sec:related}

We have discussed a broad body of related work throughout the paper. In this section, we focus on the strands most closely related to our framework and clarify its relationship to existing approaches.

\mypara{EHR foundation models.}
Structured EHR encoders such as BEHRT~\citep{li2020behrt}, Med-BERT~\citep{rasmy2021medbert},
CEHR-BERT~\citep{pang2021cehrbert} and CEHR-GPT~\citep{pang2024cehrgpt}
model the longitudinal records as patient sequences and pretrain with masked or autoregressive objectives. These models perform well on many downstream tasks, but none treat cohort structure as a first-class modelling objective or output.
EHRSHOT~\citep{wornow2023ehrshot} probes their few-shot capabilities, and MedHELM~\citep{bedi2025medhelm} broadens the evaluation scope.
Further, the review~\citep{wornow2023shaky} provides a systematic discussion of their current
limitations. 
Broader surveys, including~\citet{shickel2018deepehr} and~\citet{krishnan2022ssl},
provide the historical and methodological context for this research line.

\mypara{Patient similarity and cohort discovery.}
The technical predecessors of cohort anchoring can be grouped into two broad research directions.
\emph{Patient similarity} methods focus on retrieving clinically similar patients through
learned representations, including time-aware LSTMs~\citep{baytas2017tlstm}, deep
similarity learning frameworks~\citep{suo2018deep}, multilevel
hierarchical embeddings~\citep{choi2018mime}, and large-scale
unsupervised patient stratification approaches~\citep{landi2020deep}.
In contrast, \emph{Cohort discovery} methods aim to identify inspectable patient groupings that can support clinical understanding and downstream decision-making. 
Representative examples include prototype
networks~\citep{ming2020protosteer, barnett2021case}, ontology-aware concept
embeddings~\citep{choi2017gram}, anchor-and-learn weakly supervised
phenotyping~\citep{halpern2016electronic}, and end-to-end cohort discovery frameworks~\citep{cai2024cohortnet, liu2025neuralcohort,
zheng2024exploiting}.
While these two directions share common infrastructure, they differ in
what is returned at inference time. 
\cafm is more closely aligned with the cohort discovery paradigm and
treats the patient similarity paradigm as a debugging mechanism (Section~\ref{sec:discussion}).

\mypara{Multimodal models and clustering.}
Recent advances in multimodal biomedical representation learning provide
important context for the proposed framework.
Representative examples include BiomedCLIP~\citep{zhang2023biomedclip}, CONCH~\citep{lu2024conch},
PLIP~\citep{huang2023plip},
Med-Flamingo~\citep{moor2023medflamingo}, approaches for visual--tabular
alignment~\citep{hager2023multimodal}, and balanced multimodal
training strategies~\citep{wang2020what, peng2022balanced}.
These methods primarily focus on learning effective cross-modal representations, whereas the contamination problem we focus on was articulated by \citet{su2025clear}.
In addition, the cluster-collapse phenomenon addressed
by \cafm through the cluster-balance term has been extensively studied in the
self-supervised clustering literature~\citep{caron2020swav, caron2021dino};
we adapt the standard remedies, including entropy-regularisation and momentum encoders, to the cohort-anchored regime.

\mypara{Retrieval and generation for clinical text.}
The text-oriented counterpart to cohort retrieval lies in the rapidly growing
literature on retrieval-augmented generation~\citep{lewis2020rag, gao2024retrievalsurvey}, which augments foundation models with external evidence at inference time. 
Within the medical domain, MedRAG~\citep{xiong2024medrag} and
RAM-EHR~\citep{xu2024ramehr} represent prominent examples of this
paradigm applied to clinical and biomedical tasks.
Complementing these studies, \citet{lievin2024canllm} examine the reasoning capabilities of large
language models on medical questions,
and the medical capabilities of several leading foundation-model families have been characterised in recent evaluation work~\citep{nori2023capabilities, saab2024medgemini,
singhal2025medpalm2}.
Relative to this literature, \cafm focuses not on
retrieving textual evidence for generation, but on retrieving clinically
meaningful patient cohorts as the primary unit of evidence for downstream
reasoning and decision support.

\mypara{Causal inference for clinical decision support.}
A related line of research studies clinical decision support through the
lens of causal inference. In particular, \citet{schulam2017reliable}
formalise reliable decision support as a counterfactual estimation problem,
treating the policy under which cohort outcomes were observed. 
We view such causal-inference machinery as
complementary to cohort anchoring and discuss the relationship between these two
perspectives in greater detail in Section~\ref{sec:discussion}.

\mypara{Federated learning and privacy in medicine.}
Federated learning has emerged as a prominent paradigm for developing
clinical AI systems across institutional boundaries while limiting the
exchange of patient-level data.
\citet{rieke2020federated} provide a comprehensive overview of this area and its applications in healthcare.
The \cafm framework we propose is compatible with federated cohort indices, where institutions share cohort modules while retaining the underlying patient records locally. 
Such a setting is particularly attractive for cohort-anchored models, as the cohort
naturally serves as an intermediate unit of exchange between individual
patients and fully aggregated population statistics.
More broadly, the development and deployment of clinical AI systems remain subject to
established privacy, governance, and ethical requirements, which \citet{gerke2020ethical} survey in the context of AI-driven healthcare.

\mypara{Trustworthy and responsible clinical AI.}
The motivation for cohort anchoring is closely aligned with a broader body of
work examining the limitations and risks of deploying AI in clinical
practice.
Foundational
critiques~\citep{ghassemi2021false, rudin2019stop, obermeyer2019dissecting, gianfrancesco2018potential, finlayson2021clinician,
raji2022fallacy, futoma2020myth} highlight associated concerns and motivate the design choices underlying \cafm. 
At the same time, emerging reporting standards such as
TRIPOD+AI~\citep{collins2024tripod} and
TRIPOD-LLM~\citep{gallifant2025tripodllm}, together with real-world deployment and prospective evaluation studies~\citep{sendak2020realworld, escobar2020automated,
tomasev2019nature},
provide a view of the methodological and translational standards toward which the field is evolving.
We view cohort anchoring as complementary to these efforts, offering an architectural perspective on how clinical AI systems may become more inspectable, accountable, and clinically actionable.

\mypara{Systems and infrastructure.}
The practical realisation of cohort-anchored foundation models depends on a
broader ecosystem of systems and infrastructure research. 
This includes distributed training platforms~\citep{ooi2015singa},
machine learning lifecycle management systems~\citep{luo2021mlcask}, 
together with healthcare data modelling methods~\citep{cai2021armnet, zheng2022dyhealth, cai2022elda, zheng2020tracer} and clinical decision-support systems~\citep{zheng2021pace, zheng2022edental}.
While these works do not directly address cohort anchoring, they constitute the system substrate that spans large-scale training, data management, model maintenance, and real-world integration, on which the proposed framework can be implemented, evaluated, and ultimately deployed in practice.

\section{Conclusion}
\label{sec:conclusion}

The next step for EHR foundation models is a structural commitment to
\emph{cohort anchoring}: the explicit, end-to-end treatment of patient cohorts as representational primitives, alignment targets, retrieval
objects, and evaluation units. 
The central argument is methodological. The
\cafm framework brings together four research directions that have largely matured in parallel (deviation-aware cohort construction, cohort-conditioned pretraining, multimodal cohort alignment, and clinician-in-the-loop refinement) within a unified
design.
This integration enables a qualitatively different deployment
paradigm: one that shifts from opaque risk scores to auditable cohort-based
evidence, and from task-specific optimisation to population-grounded
foundations that can support a broad range of clinical applications across
healthcare systems.

Three claims summarise this perspective. 
First, the cohort, rather than the individual prediction, should serve as the primary unit of clinical
reasoning. Foundation-model pipelines that explicitly represent and operate
on cohorts are therefore better positioned to support deployment, auditing,
and continual updating in clinical settings.
Second, cohort anchoring is methodologically practical. Its four stages build upon existing techniques, and the framework retrofits onto contemporary EHR backbones without
requiring the underlying encoder architecture to be redesigned. As a
result, adoption primarily involves engineering and integration effort
rather than the development of entirely new modelling paradigms.
Third, the framework is empirically testable. The hypotheses outlined in
Section~\ref{sec:hypotheses} translate the proposed perspective into concrete, falsifiable claims,
allowing cohort anchoring to be evaluated against objective empirical criteria rather than relying on conceptual or rhetorical appeal alone.

Operationally, the most practical path to adoption is to augment an existing
pretrained EHR backbone with a cohort module and release the resulting
cohort assignments alongside the model card and deployment documentation.
The benefits of auditing emerge immediately: once a cohort retrieval index
is constructed, cohort summaries become directly inspectable, providing a
transparent view of the populations that underpin model behaviour.
Furthermore, deployment reports can move beyond aggregate performance
metrics to include cohort-resolved evidence, which we argue should become a
standard component of future benchmark and evaluation protocols.
In our experience, the primary challenge is not the encoder architecture itself,
but the organisational process required to curate, validate, and maintain
cohort definitions as clinical populations and practices evolve over
time.

Realising this vision requires a community-wide commitment to treating
cohorts as first-class objects that are designed, evaluated, documented, and
maintained alongside the models that learn from them. 
The next phase of progress in EHR foundation models may therefore come less from developing
new backbone architectures and more from equipping existing models with the
population-grounded substrate on which clinical reasoning has traditionally
relied. Equally important is the systems and organisational effort required
to integrate this substrate into routine clinical and institutional
workflows.
From this perspective, the principal barriers to cohort anchoring
are not conceptual. The necessary modelling components already exist. The
remaining challenge lies in engineering, deployment, governance, and the
long-term maintenance required to translate the framework from a research
proposal into a sustainable clinical capability.

\balance
\bibliography{references}

\end{document}